\documentclass[sigconf, nonacm]{acmart}

\usepackage{amsmath,amssymb,amsfonts}
\usepackage{algorithmic}
\usepackage{graphicx}
\usepackage{textcomp}
\usepackage{xcolor}
\usepackage{subfigure}
\usepackage{url}
\usepackage{comment}
\usepackage{booktabs}
\usepackage{paralist}
\usepackage{balance}

\newcommand\vldbdoi{XX.XX/XXX.XX}
\newcommand\vldbpages{XXX-XXX}
\newcommand\vldbvolume{14}
\newcommand\vldbissue{1}
\newcommand\vldbyear{2020}
\newcommand\vldbauthors{\authors}
\newcommand\vldbtitle{\shorttitle}
\newcommand\vldbavailabilityurl{https://github.com/zjs123/Booster}
\newcommand\vldbpagestyle{plain}

\begin{document}
\title{Towards Pattern-aware Data Augmentation for Temporal Knowledge Graph Completion}

\author{Jiasheng Zhang}
\affiliation{%
  \institution{University of Electronic Science and Technology of China}
  \city{Chengdu}
  \country{China}}
\email{zjss12358@std.uestc.edu.cn}

\author{Deqiang Ouyang}
\affiliation{%
  \institution{Chongqing University}
  \city{Chongqing}
  \country{China}}
\email{deqiangouyang@cqu.edu.cn}

\author{Shuang Liang}
\author{Jie Shao}
\affiliation{%
  \institution{University of Electronic Science and Technology of China}
  \city{Chengdu}
  \country{China}}
\email{{shuangliang,shaojie}@uestc.edu.cn} 
\begin{abstract}
Predicting missing facts for temporal knowledge graphs (TKGs) is a
fundamental task, called temporal knowledge graph completion (TKGC).
One key challenge in this task is the imbalance in data
distribution, where facts are unevenly spread across entities and
timestamps. This imbalance can lead to poor completion performance
for long-tail entities and timestamps, and unstable training due to
the introduction of false negative samples. Unfortunately, few
previous studies have investigated how to mitigate these effects.
Moreover, for the first time, we found that existing methods suffer
from model preferences, revealing that entities with specific
properties (e.g., recently active) are favored by different models.
Such preferences will lead to error accumulation and further
exacerbate the effects of imbalanced data distribution, but are
overlooked by previous studies. To alleviate the impacts of
imbalanced data and model preferences, we introduce
\textit{Booster}, the first data augmentation strategy for TKGs. The
unique requirements here lie in generating new samples that fit the
complex semantic and temporal patterns within TKGs, and identifying
hard-learning samples specific to models. Therefore, we propose a
hierarchical scoring algorithm based on triadic closures within
TKGs. By incorporating both global semantic patterns and local
time-aware structures, the algorithm enables pattern-aware
validation for new samples. Meanwhile, we propose a two-stage
training approach to identify samples that deviate from the model's
preferred patterns. With a well-designed frequency-based filtering
strategy, this approach also helps to avoid the misleading of false
negatives. Experiments justify that \textit{Booster} can seamlessly
adapt to existing TKGC models and achieve up to an 8.7\% performance
improvement.
\end{abstract}

\maketitle

\pagestyle{\vldbpagestyle}
\begingroup\small\noindent\raggedright\textbf{PVLDB Reference Format:}\\
\vldbauthors. \vldbtitle. PVLDB, \vldbvolume(\vldbissue): \vldbpages, \vldbyear.\\
\href{https://doi.org/\vldbdoi}{doi:\vldbdoi}
\endgroup
\begingroup
\renewcommand\thefootnote{}\footnote{\noindent
This work is licensed under the Creative Commons BY-NC-ND 4.0 International License. Visit \url{https://creativecommons.org/licenses/by-nc-nd/4.0/} to view a copy of this license. For any use beyond those covered by this license, obtain permission by emailing \href{mailto:info@vldb.org}{info@vldb.org}. Copyright is held by the owner/author(s). Publication rights licensed to the VLDB Endowment. \\
\raggedright Proceedings of the VLDB Endowment, Vol. \vldbvolume, No. \vldbissue\ %
ISSN 2150-8097. \\
\href{https://doi.org/\vldbdoi}{doi:\vldbdoi} \\
}\addtocounter{footnote}{-1}\endgroup

\ifdefempty{\vldbavailabilityurl}{}{
\vspace{.3cm}
\begingroup\small\noindent\raggedright\textbf{PVLDB Artifact Availability:}\\
The source code, data, and/or other artifacts have been made available at \url{\vldbavailabilityurl}.
\endgroup
}

\section{Introduction}

\begin{figure}[t]
  \centering
  \includegraphics[width=0.72\linewidth]{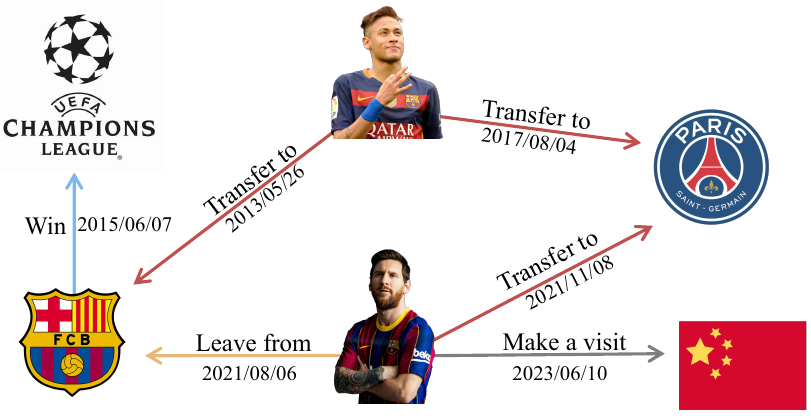}
  \caption{An illustration of temporal knowledge graph.}
  \label{fig:TKG}
\end{figure}

Temporal knowledge graphs (TKGs) are knowledge base systems that
organize dynamic human knowledge in a structured manner. They are
highly valuable for many applications such as event prediction
\cite{DBLP:journals/tois/TangCSL24} and recommendation systems
\cite{DBLP:journals/tois/ZhaoWCWTHX23}. As illustrated in
Figure~\ref{fig:TKG}, a temporal knowledge graph is a dynamic
directed graph characterized by node and edge categories, where
nodes represent entities in the real world and labeled edges signify
the relations between these entities. Each edge with its connected
nodes can form a tuple $(s,r,o,t)$ to describe a piece of dynamic
knowledge (fact) in the real world, such as $(Messi, Transfer to,
PSG, 2021/11/8)$.

Due to delays in updates and limitations of extraction tools
\cite{DBLP:journals/csur/NasarJM21, DBLP:conf/icde/ZhuWXWJWMZ23,
DBLP:conf/emnlp/TangFZ19}, TKGs are often incomplete, missing some
facts existing in reality. To address this, predicting these missing
facts---known as temporal knowledge graph completion (TKGC)
\cite{DBLP:journals/corr/abs-2308-02457}---has become a fundamental
task to improve the quality of TKGs and support downstream tasks
\cite{DBLP:journals/pvldb/GuptaSGGZLL14}. Several methods have been
developed for TKGC, falling into two categories based on how they
model temporal information: timestamp embedding models
\cite{DBLP:conf/emnlp/DasguptaRT18, DBLP:conf/coling/XuNAYL20,
DBLP:conf/iclr/LacroixOU20} that learn representation for each
timestamp, and dynamic embedding models
\cite{DBLP:conf/emnlp/WuCCH20, DBLP:conf/semweb/XuNAYL20,
DBLP:conf/aaai/GoelKBP20} that learn evolving entity and relation
representations.

Despite their effectiveness, recent studies have shown that TKGs
suffer from imbalanced data distribution
\cite{DBLP:conf/emnlp/WuCCH20}, which may seriously impair the
performance of TKG completion. Unfortunately, most existing methods
overlook this aspect. They only report performance improvement on
several metrics (e.g., mean reciprocal rank (MRR)) without
thoroughly analyzing how the imbalanced data impacts their
performance and how to alleviate such impacts, leading to less
convincing and unsatisfactory results.

\textbf{Previous limitations.} The number of facts varies
significantly across entities and timestamps. While a few have
extensive fact descriptions, most only have a handful of related
facts, resulting in imbalanced data distribution within TKGs. We
first revisit the completion performance of existing methods and
find that they are severely affected by data imbalance in two key
ways: \textbf{1) Unstable training.} Most existing methods are
trained contrastively \cite{DBLP:journals/pami/XieXZWJ23}, treating
facts existing in TKGs as positive and all others as negative. The
models then learn to distinguish between these positive and negative
samples. However, due to the imbalanced data distribution, many
valid facts are missing in TKGs and mistakenly treated as negative
samples (i.e., false negatives). This will mislead the training
process and make the completion performance of entities with few
facts deteriorate during training. \textbf{2) Imbalanced
performance.} The uneven distribution of facts across timestamps
causes the performance of existing methods to vary dramatically
depending on the timestamp, even between adjacent timestamps. The
neglect of uneven distribution makes the training process of these
methods further aggravate such performance imbalance
(Section~\ref{sec:imbalance_data_observation}).

While examining the effects of imbalanced data distribution, we
further identified a previously unrecognized problem: existing
methods also exhibit a \textbf{model preference issue} that can
worsen the impact of data imbalance. During completion, existing
methods favor entities with specific properties according to their
architectures (Section~\ref{sec:model_preference_observation}). For
example, tensor factorization-based models
\cite{DBLP:conf/iclr/LacroixOU20} prioritize frequently interacted
entities, whereas recurrent neural network-based models
\cite{DBLP:conf/emnlp/WuCCH20} favor recently active entities. Such
preferences can make it difficult for these models to learn TKG
patterns that deviate from the model's preferences, especially when
valid samples fitting these patterns are mistakenly treated as false
negatives.

\textbf{Technical challenges.} Although some data augmentation
approaches have been proposed to solve similar imbalanced data
issues for general graphs and static knowledge graphs
\cite{DBLP:journals/sigkdd/DingXTL22, DBLP:conf/ijcai/TangPZZZH022,
DBLP:conf/cikm/Manchanda23}, they face key challenges when applied
to TKGs: \textbf{1) False negative filtering.} Some of these methods
simply filter neighboring nodes as false negatives
\cite{DBLP:journals/vldb/ZhangY021}. They cannot consider various
components (i.e., entities, relations, and timestamps) within TKGs'
complex graph structure, failing to achieve adequate filtering.
\textbf{2) New samples generation.} They generate new samples solely
based on node connectivity. However, TKGs have intricate semantic
and temporal patterns brought by diverse entity and relation
categories, as well as time-evolving topology. New samples must
therefore fit with these patterns. \textbf{3) Training procedure.}
Most of them simply train the models on the refined graph data
without considering model preference issues, leading to low
generalization to various TKG patterns.

\textbf{The proposed work.} We present \textit{Booster}, the first
pattern-aware data augmentation strategy specialized to TKGs to
tackle their imbalanced data and model preference issues. It uses
three frequency-based filtering strategies tailored to different
components of TKGs, considering both intra-component and
inter-component interaction frequencies to adequately filter
potential false negatives. A hierarchical scoring algorithm then
classifies these samples as either hard negatives or false
negatives, ensuring the identified false negatives fit both the
global semantic patterns shared across all facts and the recent
trends within the local graph structure, which can be used to enrich
the imbalanced data. Finally, a two-stage training approach is
proposed. The models are first pre-trained on filtered high-quality
facts to avoid the misleading of false negatives and identify
preference-deviated facts, and then fine-tuned on hard negatives,
false negatives, and preference-deviated facts to inject
pattern-aware fine-grained information while enhancing pattern
generalization ability. Experiments on 5 real-world TKGs show that
\textit{Booster} can seamlessly adapt to existing TKGC methods,
improving their performance up to 8.7\%, surpassing typical temporal
graph and knowledge graph data augmentation techniques on average of
7.1\%, while also reducing performance variance of existing TKGC
models by 22.8\% on average. Our contributions are as follows:
\begin{compactitem}
    \item We make the first attempt to investigate the imbalanced data and
model preference issues of TKG completions.
    \item We propose the first pattern-aware data augmentation strategy
tailored to TKGs---\textit{Booster}, which can generate new samples
fitting TKG patterns and enhance the model's generalization ability
to different patterns.
    \item Experimental results show that \textit{Booster} can effectively
improve the performance of existing TKGC models.
\end{compactitem}

\section{Related Work}

\subsection{Temporal Knowledge Graph Completion}

Temporal knowledge graph completion aims to predict missing facts
based on observed ones within TKGs. Existing methods can be divided
into two categories: 1) Timestamp embedding methods
\cite{DBLP:journals/isci/ZhangCSCL24, DBLP:journals/eswa/YangYSX24,
DBLP:conf/iclr/LacroixOU20} that learn representations respectively
for entities, relations, and timestamps, and use these embeddings to
predict the missing facts. For example, HyTE
\cite{DBLP:conf/emnlp/DasguptaRT18} integrates the learnable
timestamp embeddings into the translation function of the TransE
model \cite{DBLP:conf/nips/BordesUGWY13}. TNT
\cite{DBLP:conf/iclr/LacroixOU20} uses 4-order tensor factorization
to generate these embeddings. On this basis, Timeplex
\cite{DBLP:conf/emnlp/JainRMC20} extends by considering the
recurrent nature of relations, while TELM
\cite{DBLP:conf/naacl/XuCNL21} extends by learning multi-vector
representations with canonical decomposition. Recently, QDN
\cite{DBLP:journals/tnn/WangWGLHY24} uses a quadruplet distributor
network to support the fourth-order factorization. MADE
\cite{DBLP:journals/tcyb/WangWGPLYG24} proposes to learn
multi-curvature representations. 2) Dynamic embedding methods
\cite{DBLP:conf/emnlp/DasguptaRT18, DBLP:conf/coling/XuNAYL20,
DBLP:conf/iclr/HanCMT21, DBLP:conf/www/ZhangX0WW23} that learn
time-evolving entity and relation representations to model their
changing semantics. DE \cite{DBLP:conf/aaai/GoelKBP20} uses
nonlinear operations to model various evolution trends of entity
semantics. TA \cite{DBLP:conf/emnlp/Garcia-DuranDN18} utilizes a
sequence model to generate time-specific relation representations.
CENET \cite{DBLP:conf/aaai/XuO0F23} employs historical contrastive
learning to learn temporal dependencies. Recently, some studies have
attempted to model structure information of TKGs via graph neural
networks \cite{DBLP:conf/icml/GilmerSRVD17}. For example, TEMP
\cite{DBLP:conf/emnlp/WuCCH20} uses self-attention to model the
spatial and temporal locality. RE-GCN
\cite{DBLP:conf/sigir/LiJLGGSWC21} auto-regressively models
historical sequence. LogCL \cite{DBLP:conf/icde/ChenWWZCLL24} learns
both local and global historical structures.

Despite their effectiveness, unfortunately, none of them deeply
investigated the imbalanced data distribution issue
\cite{DBLP:conf/emnlp/WuCCH20} inherent in the TKGs. Neither how
imbalanced data affect their completion performance, nor how to
alleviate such impacts are studied, making their results less
convincing and unsatisfactory. A recent work, TILP
\cite{DBLP:conf/iclr/XiongYFK23}, claims that its logic rule-based
method is less affected by imbalance. However, its strategy lacks
the adaptability to other TKGC models.

\subsection{Graph Data Augmentation}

Some data augmentation strategies have been developed recently to
improve the data quality for graphs \cite{DBLP:conf/icml/HanJLH22,
DBLP:conf/icde/CuiCYDF024, DBLP:conf/cikm/Manchanda23,
DBLP:journals/tkde/LuZGLFWZ24, DBLP:conf/icde/ZhangXCY0J24}, which
helps to reduce the effects of imbalanced data distributions in
general graph modeling (e.g., degree bias
\cite{DBLP:conf/cikm/TangYSWTAMW20}). For example, AIA
\cite{DBLP:conf/nips/SuiWWCLZW023} adversarially generates masks on
graphs to handle distribution shift. GraphPatcher
\cite{DBLP:conf/nips/Ju00S023} generates virtual nodes for
ego-graphs to mitigate the degree bias. Recently, some studies have
also explored the data augmentation strategies for temporal graphs
\cite{DBLP:conf/kdd/TianJHGQ24}. For example, MeTA
\cite{DBLP:conf/nips/WangCLDWBH21} modifies temporal topology and
features to enhance model robustness. TGEditor
\cite{DBLP:conf/icaif/ZhangZ023} conducts task-guided graph editing
for temporal transaction networks. However, these strategies are
designed for general graphs without considering complex semantics
brought by node and edge categories, and thus fail to generate
samples that fit TKG patterns.

Recently, the issue of imbalanced data in static knowledge graphs
has also gained increasing attention. Some studies develop advanced
negative sampling strategies to avoid false negatives
\cite{DBLP:conf/icml/KamigaitoH22,
DBLP:journals/corr/abs-2402-19195, DBLP:conf/semweb/YaoLYLB23}. For
example, NSCaching \cite{DBLP:journals/vldb/ZhangY021} uses an
importance sampling approach to adaptively identify high-quality
negative samples. DeMix \cite{DBLP:conf/semweb/ChenZYCT23}
introduces a self-supervised mechanism to identify negative samples.
However, these strategies encounter challenges when applied to TKGs,
as they fail to consider the temporality of facts. Moreover, these
methods often rely on path searching or adversarial training, which
can be highly time-consuming for TKGs with lengthy historical
sequences. Some researches also investigate to generate new facts to
enrich the imbalanced data \cite{DBLP:conf/ijcai/TangPZZZH022,
DBLP:conf/cikm/Manchanda23, DBLP:conf/coling/Yao0X022}. For example,
KG-Mixup \cite{DBLP:conf/www/ShomerJ0T23} generates synthetic facts
in the embedding space to mitigate degree bias. KGCF
\cite{DBLP:conf/www/ChangCL23} augments with counterfactual
relations. However, they can neither generate new facts that fit the
complex semantic and temporal patterns within TKGs, nor do they
address model preference issues.

Overall, although data augmentation has been widely studied to
address data imbalance for general graph learning, it remains an
unexplored area for TKGs. Existing data augmentation strategies
either face high time-consuming for TKGs with lengthy historical
sequences, or are unaware of complex semantic and temporal patterns
within TKGs, leading to an urgent for a data augmentation strategy
specialized to TKGs.

\begin{figure*}[t]
\centering
\subfigure[]{\includegraphics[width=0.3\linewidth]{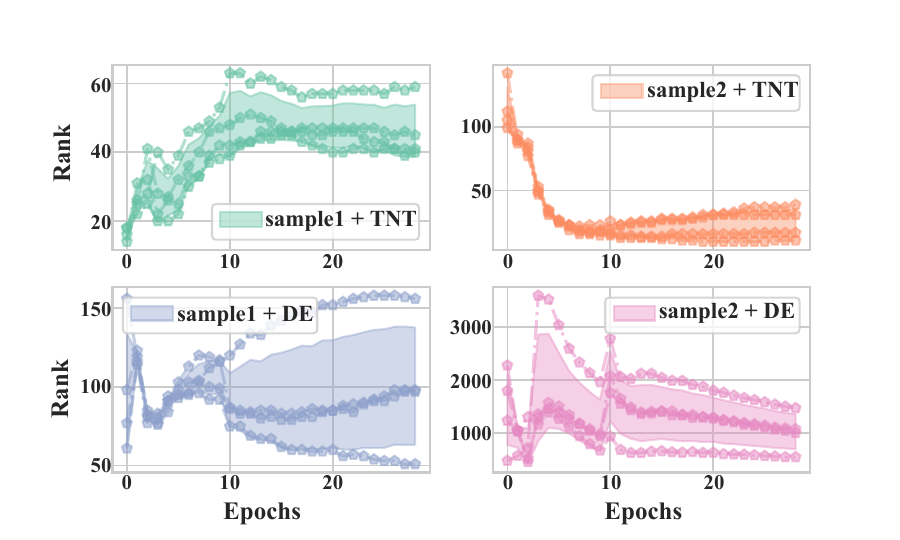}}
\subfigure[]{\includegraphics[width=0.3\linewidth]{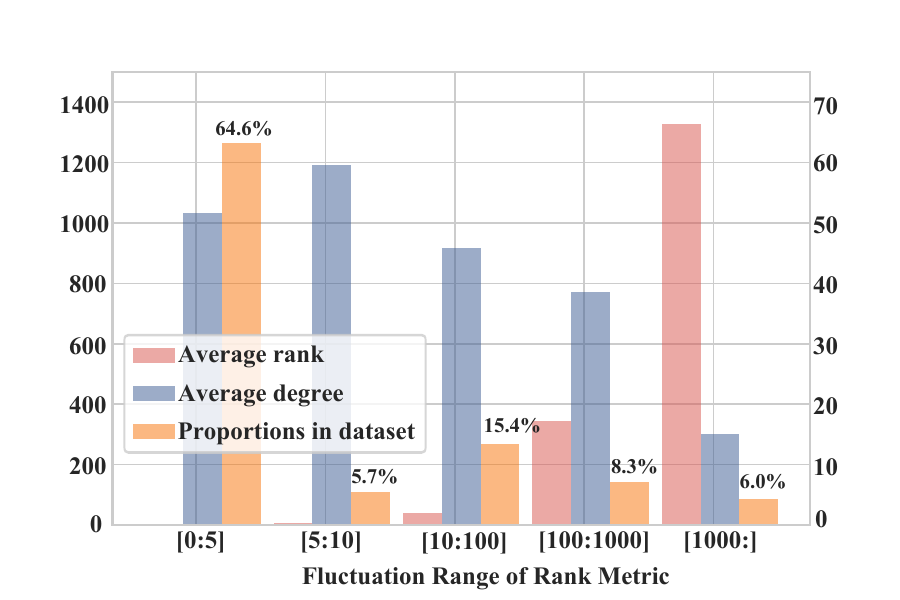}}
\subfigure[]{\includegraphics[width=0.3\linewidth]{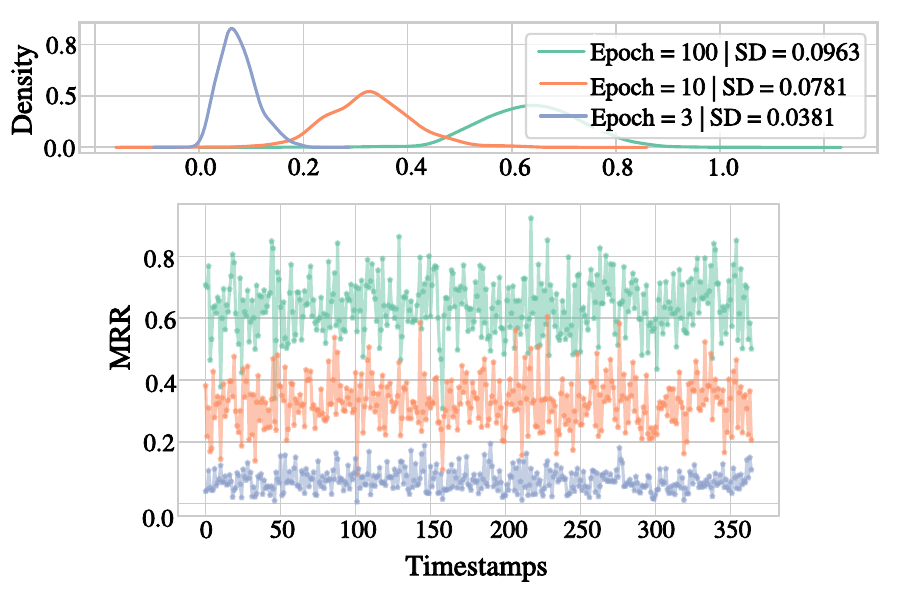}}
\subfigure[]{\includegraphics[width=0.3\linewidth]{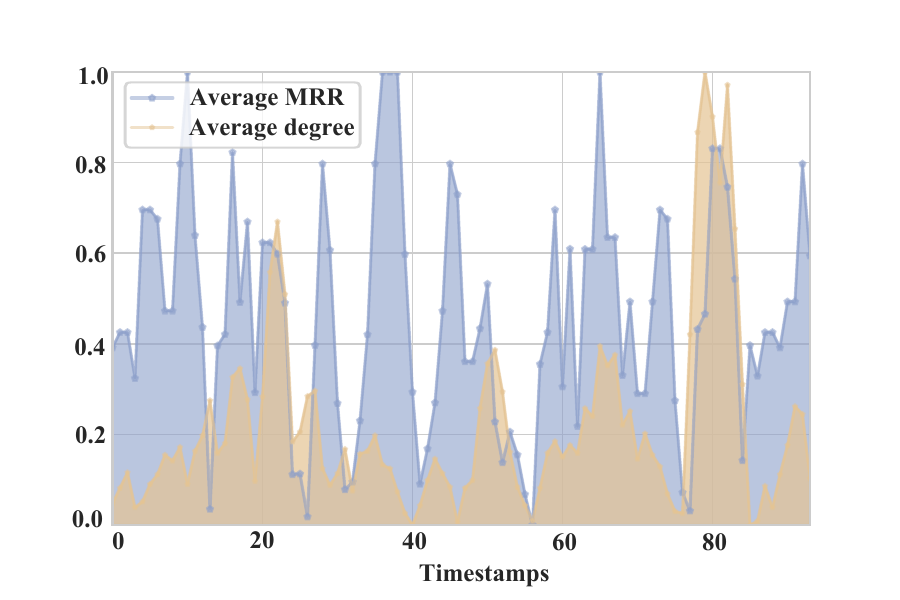}}
\subfigure[]{\includegraphics[width=0.3\linewidth]{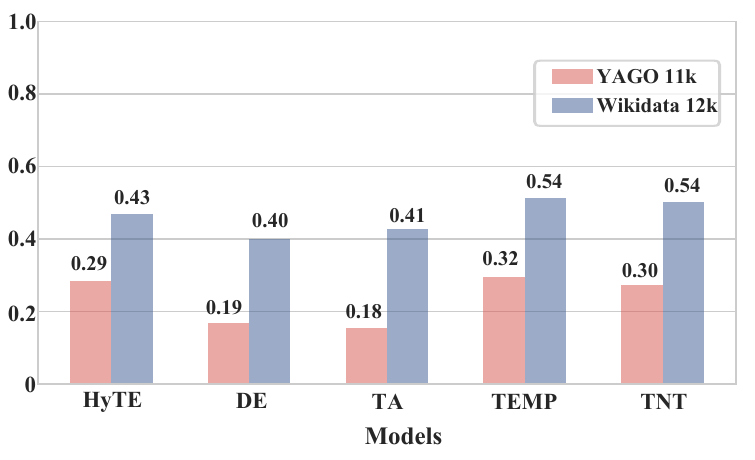}}
\subfigure[]{\includegraphics[width=0.3\linewidth]{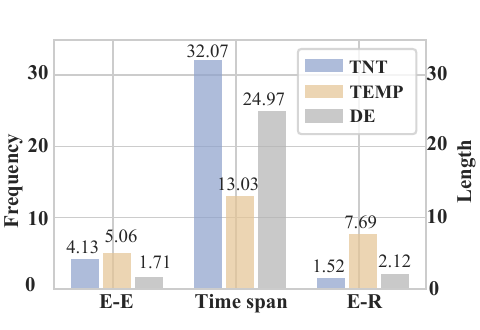}}
\caption{(a) Change of the $rank$ metric during training. Each
training is independently repeated four times and the color-filled
part is the fluctuation range of the $rank$ metric across four
training runs. (b) The average degree of samples with different
$rank$ fluctuation ranges. (c) MRR of the TEMP model across
different timestamps. The top plot displays the density function of
the MRR distribution at different epochs. SD refers to standard
deviation. (d) MRR and the average degree of entities in different
timestamps. (e) Proportions of positive samples among the top-10
ranked candidates for different models. (f) The statistical
characteristics of the top-ranked entities for different models.}
\label{fig:observations}
\end{figure*}

\section{Preliminary Study}

\subsection{Temporal Knowledge Graph}

A temporal knowledge graph is denoted as $\mathcal{G} =
(\mathcal{E}, \mathcal{R}, \mathcal{T}, \mathcal{F})$. $\mathcal{E}$
and $\mathcal{R}$ are entity set and relation set, respectively.
$\mathcal{T}$ is the set of observed timestamps and $\mathcal{F}$ is
the set of facts. Each tuple $(s, r, o, t) \in \mathcal{F}$ connects
the subject and object entities $s, o \in \mathcal{E}$ via a
relation $r \in \mathcal{R}$ in timestamp $t \in \mathcal{T}$, which
means a unit knowledge (i.e., a fact).

\subsection{Temporal Knowledge Graph Completion}

Temporal knowledge graph completion aims to predict the missing
facts through the existing ones. Given an incomplete fact $(s, r, ?,
t)$, the task identifies the most likely object entity $o_c$ from
the candidate set $\mathcal{E}$. Each candidate fact $(s, r, o_c,
t)$ is ranked by confidence score, and the highest-ranking candidate
is chosen as the new fact. The rank of the true object entity,
denoted as $rank(s,r,o,t)$, is the basic metric of this task (lower
is better). It indicates the position of the correct object entity
$o$ among all candidate entities $o_c \in \mathcal{E}$. Building on
this, mean reciprocal rank (MRR) is calculated as the average
reciprocal rank across all facts, defined as:
\begin{equation} \small
    MRR = \frac{1}{|Test|} \sum_{(s,r,o,t) \in Test} \frac{1}{rank(s,r,o,t)}.
\end{equation}
A higher MRR value indicates better model performance.

\subsection{Imbalanced Data Distribution}
\label{sec:imbalance_data_observation}

In this part, we selected three typical TKGC methods (i.e., DE
\cite{DBLP:conf/aaai/GoelKBP20}, TNT
\cite{DBLP:conf/iclr/LacroixOU20}, and TEMP
\cite{DBLP:conf/emnlp/WuCCH20}) to evaluate their completion
performance on the most popular TKG dataset ICEWS14 \cite{ICEWS}.
Our results demonstrate issues with unstable training and
inconsistent performance. Further analysis shows that these
limitations stem from the imbalanced data distribution within TKGs,
which for the first time reveals the impacts of imbalanced data
distribution on existing TKGC methods.

\textbf{Unstable training.} As shown in
Figure~\ref{fig:observations}(a), existing methods suffer from
unstable training in two aspects: 1) The effectiveness of training
is inconsistent across different samples. The $rank$ metrics of some
facts may gradually increase during training (e.g., sample 1 in the
TNT model), showing that training can unintentionally damage their
completion performance. Meanwhile, the $rank$ metrics for some facts
can only fluctuate near a large number (e.g., sample 2 in the DE
model), indicating that training cannot effectively optimize their
completion performance. 2) The effectiveness of training is
inconsistent across different runs. In the DE model, sample 1 shows
highly inconsistent performance across four independent training
runs. The fluctuation range of the $rank$ metric widens as the
training progresses, indicating that the inconsistency becomes more
serious.

To study the cause of the unstable training, we split the facts into
different sets based on the fluctuation range of the $rank$ metric
across four independent runs, and then calculate the average degree
for entities involved in each set. As shown in
Figure~\ref{fig:observations}(b), we can see that as the average
degree decreases, both the fluctuation range and the average $rank$
increase, highlighting that unstable training is more likely to
occur for entities with sparse local structures. This is because
existing methods typically focus on distinguishing between positive
and negative samples, treating facts in TKGs as positive and all
others as negative. However, entities with sparse local structures
often have missing yet valid facts, introducing many false negative
samples that will mislead the training
\cite{DBLP:journals/vldb/ZhangY021}.

\textbf{Imbalanced performance.} As shown in
Figure~\ref{fig:observations}(c), existing methods suffer from
imbalanced performance: 1) MRR varies significantly across different
timestamps, with considerable gaps even between adjacent timestamps.
2) The performance can be notably poor for some timestamps. 3) As
training progresses, the standard deviation of MRR will increase,
suggesting that the performance imbalance of existing methods may
worsen over training. To investigate the cause of imbalanced
performance, as shown in Figure~\ref{fig:observations}(d), we
calculate the average degree and the average MRR for each timestamp.
We can see that there exists a synchronization between the
fluctuation trends of the average degree and MRR, with their peaks
and troughs occurring simultaneously, albeit with some displacement
and scaling. This finding suggests that an imbalanced data
distribution across timestamps can lead to performance imbalance for
existing TKGC methods.

These observations suggest that current TKGC methods face
significant challenges due to unstable training and imbalanced
performance stemming from imbalanced data distribution. Addressing
these issues is crucial to improving their effectiveness.

\subsection{Model Preference}
\label{sec:model_preference_observation}

Some studies use self-training to address data imbalance, selecting
high-scoring unlabeled samples as pseudo-positive samples and adding
them to the training data \cite{DBLP:conf/www/LiuHWSZZ22}. However,
does this strategy also work for temporal knowledge graph completion
models? As shown in Figure~\ref{fig:observations}(e), the top-ranked
samples of existing TKGC models are not always positive (e.g., only
30\% of top-10 ranked samples by TNT are positive), indicating that
using self-learning for these TKGC models will introduce a large
number of noisy samples. Although reducing the size of
pseudo-positive samples can help mitigate such noise, the
challenging issue of model preference continues to hinder
self-training in TKGC models---an aspect overlooked in previous
research.

As shown in Figure~\ref{fig:observations}(f), we analyze the data
properties of the top-ranked entities from different models. ``E-E"
means the frequency of the candidate entity interacting with other
entities. ``Time span" means the length between the test timestamp
and the candidate entity's nearest active timestamp. ``E-R" means
the frequency of the candidate entity interacting with the relation
in the query. We can see that the top-ranked entities from different
models have significant differences in their data properties. For
example, TNT prefers frequently interacted entities, while TEMP
prefers recently active entities and entities that have interactions
with the query relation, revealing that existing TKGC methods suffer
from \textbf{model preference} that they prefer entities with
specific properties. Previous studies have shown that facts in TKGs
often follow diverse patterns \cite{DBLP:conf/icde/XinC24}. However,
the model preference issue makes TKGC models struggle to learn facts
that deviate from their favored patterns (especially when they are
mistakenly treated as false negatives). This challenge becomes even
more serious with self-training, as the model consistently selects
samples with one pattern as new training data, which degrades the
model's generalization ability to other TKG patterns.

These observations show that current TKGC methods suffer from model
preferences, limiting the effectiveness of traditional data
augmentation strategies (e.g., self-training). Unfortunately, no
previous work has investigated how to address this challenge.

\section{Method}

\begin{figure}[t]
  \centering
  \includegraphics[width=1\linewidth]{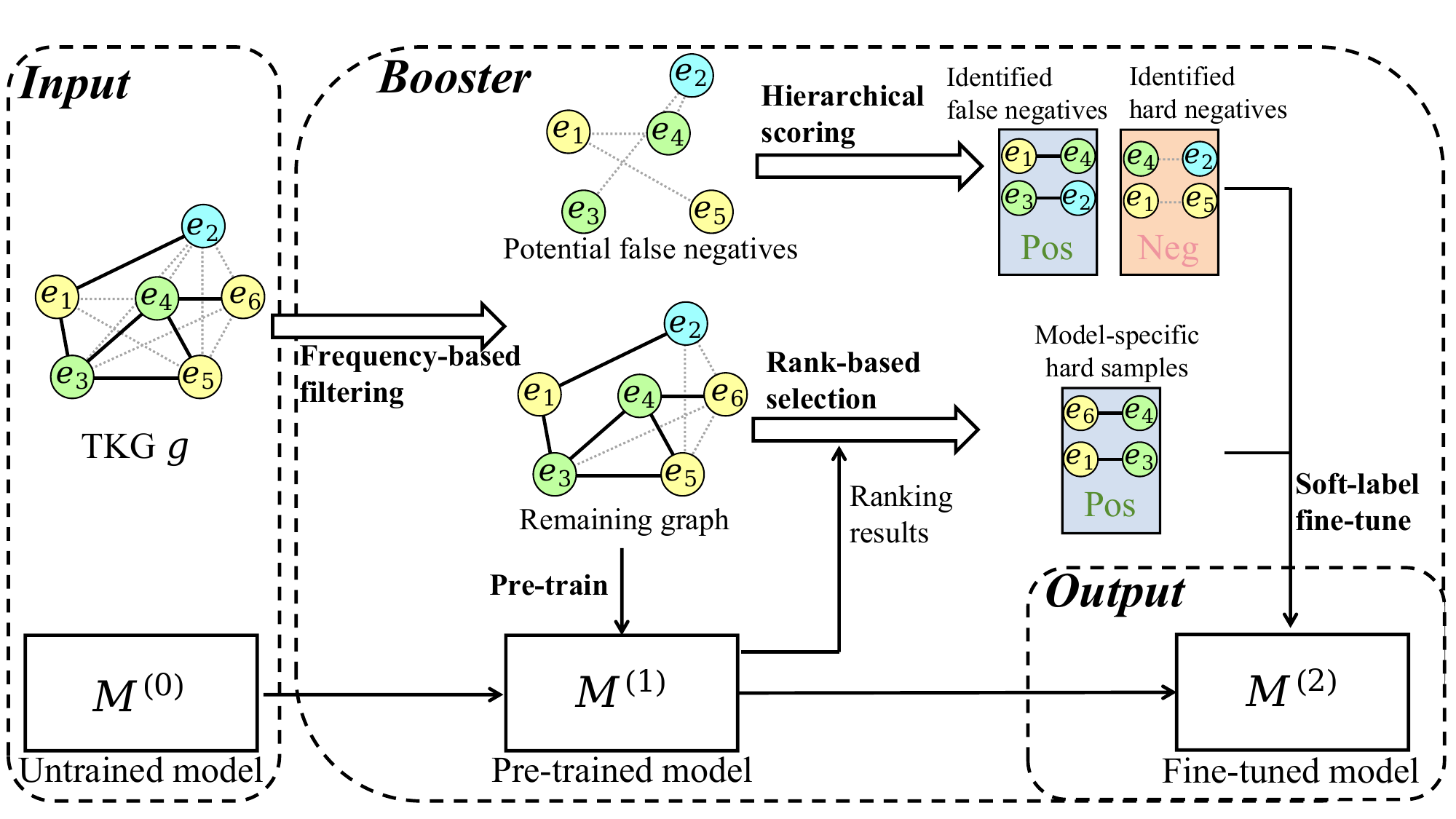}
  \caption{The conceptual illustration of the overall architecture
of \textit{Booster}, where the black solid lines indicate the
observed facts in TKG and the gray dashed lines indicate facts do
not exist in TKG. We hide the time annotations and edge types in the
figure for brevity.}
  \label{fig:booster}
\end{figure}

\subsection{Overall Architecture}

Our above discussions emphasize the urgent need for a data
augmentation strategy tailored to TKGs, enriching imbalanced data
distribution, and alleviating misleading from false negatives and
model preferences. To this end, we introduce \textit{Booster}, a
data augmentation framework that can generate pattern-aware new
samples to enrich TKG structure and reduce misleading through a
two-stage training approach.

As shown in Figure~\ref{fig:booster}, \textit{Booster} takes a TKG
$\mathcal{G}$ and an untrained TKGC model $M^{(0)}$ as input. In the
first stage, it uses a frequency-based filtering strategy to filter
out potential false negatives, and then pre-trains $M^{(0)}$ on the
remaining data. This step serves a main purpose: protecting the
model from misleading false negatives, and thus enabling accurate
selection of preference-deviated samples. Low-ranked positive
samples reveal which facts the model is hard to learn (i.e., those
it does not prefer).

In the second stage, a hierarchical scoring algorithm is used to
further separate potential false negatives into real false negatives
and hard negatives. The pre-trained model $M^{(1)}$ is then
fine-tuned on these identified samples, along with
preference-deviated facts, to produce the final model $M^{(2)}$.
This process has two key benefits: 1) enriching sparse data by
incorporating validated false negatives as additional positive
samples, and 2) emphasizing hard samples unique to the data and
model to reduce misleading stemming from data imbalances and model
preferences. Notably, \textit{Booster} is plug-and-play and
compatible with any existing TKGC model.

\subsection{Frequency-based Filtering}
\label{sec:frequency-based_filtering}

Compared with general graph data, filtering potential false
negatives for TKGs faces several unique challenges: 1) TKGs contain
various components (i.e., entities, relations, and timestamps). The
co-occurring patterns within TKGs often involve several different
components, requiring the filtering strategy to fully consider these
intra- and inter-component patterns. 2) The integration of time
annotations results in a long historical graph sequence and multiple
edges between nodes, requiring the filtering strategy to consider
the distribution of these edges in the time dimension. Therefore, we
propose a frequency-based filtering approach, using tailored
strategies for each component to filter false negatives effectively.

\textbf{Relation-based filtering.} Relations in TKGs have
significant co-occurring patterns
\cite{DBLP:journals/tois/ZhaoWCWTHX23}. For example, ``economic
sanctions" and ``export restriction" are a pair of relations that
often co-occur between two hostile countries. After the relation
``transfer to" occurs between a player and a football club, the
relation ``play for" will subsequently occur between them. This
inspires us that edges missing in a TKG but fitting these relation
patterns are likely to be false negatives, and we can detect them by
identifying these patterns. Therefore, for each relation $r$ we
construct its co-occurred relation set as $R(r) = \{r_i | (s_j, r,
o_j, t_j) \in \mathcal{G}, (s_j, r_i, o_j, t'_j) \in \mathcal{G},
|t_j - t'_j| < L_r \}$ where $L_r$ is a hyper-parameter. Then, for
each observed fact $(s, r, o, t) \in \mathcal{G}$ we can filter its
corresponding potential false negatives as $(s, r_i, o, t) \notin
\mathcal{G}$ where $r_i \in R(r)$. We further refine the filtering
by considering the inter-component patterns and pattern frequencies.
First, entities have preferences to interact with a specific set of
relations (e.g., athletes are more likely to have relation ``play
for"), so recognizing these entity-relation interaction patterns
helps exclude unrealistic combinations. Second, the higher frequency
of relations within $R(r)$ indicates a more important pattern. By
retaining only the top-$m$ most frequently co-occurring relations in
$R(r)$, we can filter out low-confidence patterns, reduce potential
false negatives, and thus lower the time required for scoring in the
next step. Therefore, we filter out the relation-based false
negatives for each fact $(s, r, o, t) \in \mathcal{G}$ as:
\begin{equation} \small
    RN_{(s,r,o,t)} = \{(s, r', o, t) | r' \in \widetilde{R}(r) \cap \widetilde{R}(s), (s, r', o, t) \notin \mathcal{G} \},
\end{equation}
where $\widetilde{R}(r)$ is the subset of $R(r)$ which only
preserves top-$m$ most frequent relations. Similarly,
$\widetilde{R}(s)$ is the frequency filtered subset of $R(s) = \{r_k
| (s, r_k, o_k, t_k) \in \mathcal{G}\}$, which preserves top-$m$
frequently interacted relations of entity $s$.

\textbf{Entity-based filtering.} While the above strategy filters
false negatives based on relation semantics, the connectivity among
entities also provides insights into the occurrence of facts.
Entities often have preferences to interact with a specific set of
entities (e.g., "Israel" and "Houthis in Yemen" frequently interact
due to ongoing conflict), inspiring us that facts fitting entity
co-occurring patterns but missing in $\mathcal{G}$ are likely to be
false negatives. Therefore, for each entity $e$ we construct its
co-occurring entity set as $N(e) = \{e_i | (e, r_i, e_i, t_i) \in
\mathcal{G}\}$, and filter the corresponding top-$m$ frequency
entities as $\widetilde{N}(e)$. Subsequently, we filter out the
entity-based false negatives for each fact $(s, r, o, t) \in
\mathcal{G}$ as:
\begin{equation} \small
    EN_{(s,r,o,t)} =\{(s, r, o', t) | o' \in \widetilde{N}(s), r \in \widetilde{R}(o'), (s, r, o', t) \notin \mathcal{G} \},
\end{equation}
indicating entity pairs that are likely to connect through the
relation $r$ but are missing in TKG. This is achieved by filtering
the entity $o'$ that frequently interacts with $s$ (i.e., $o' \in
\widetilde{N}(s)$), and has prior interactions with relation $r$
(i.e., $r \in \widetilde{R}(o')$).

\textbf{Time-based filtering.} Some facts may be repeated many times
over a short period, such as ``(Region A, Armed attack, Region B)"
or ``(Country A, Hold negotiations with, Country B)". Due to the
limitation of the update frequency of TKGs, repeated facts may be
missing in some timestamps. Therefore, we can detect potential false
negatives by finding the omissive timestamp within the time interval
where the fact repeats. Specifically, we filter out the time-based
false negatives for each fact $(s, r, o, t) \in \mathcal{G}$ as:
\begin{equation} \small
    TN_{(s, r, o, t)} = \{(s, r, o, t') | t' \in [x, t], (s, r, o, x) \in \mathcal{G}, (s, r, o, t') \notin \mathcal{G}\}.
\end{equation}
Notably, we restrict that $t-x < L_t$ to ensure only focusing on
short-period repetitions, where $t-x$ indicates the fact repetition
period and $L_t$ is a hyper-parameter with a small value.

\begin{figure}[t]
  \centering
  \subfigure[ICEWS 14]{\includegraphics[width=0.35\linewidth]{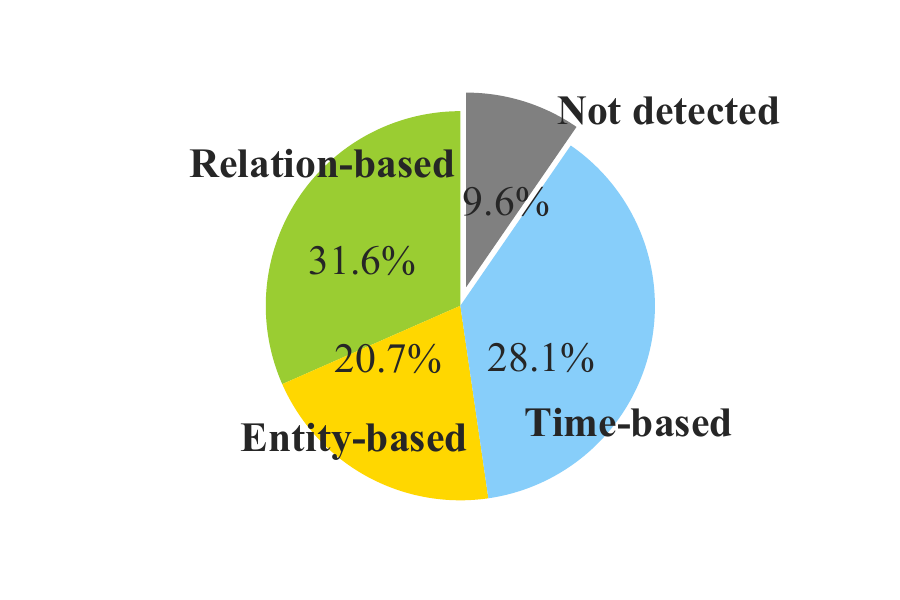}}
  \subfigure[ICEWS 05-15]{\includegraphics[width=0.35\linewidth]{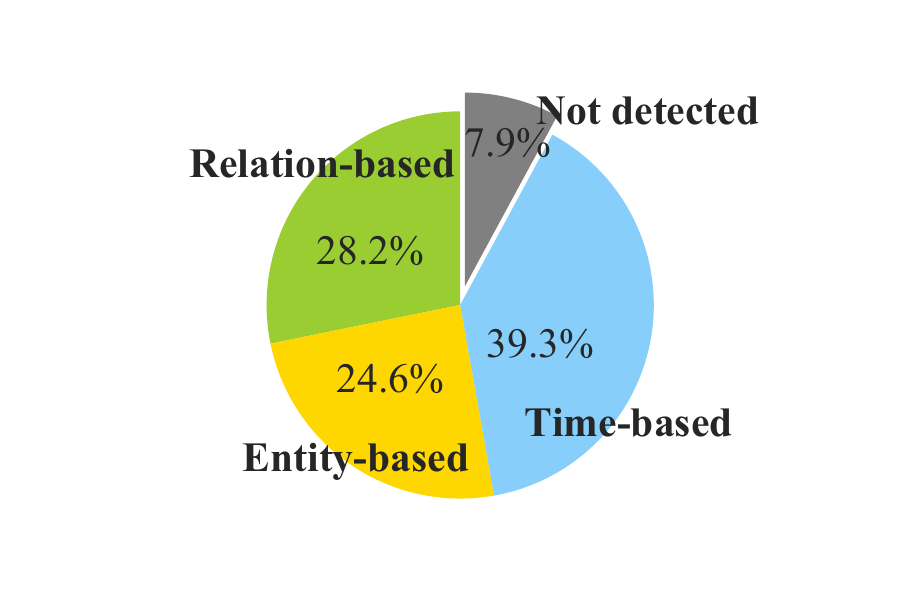}}
  \subfigure[YAGO 11k]{\includegraphics[width=0.35\linewidth]{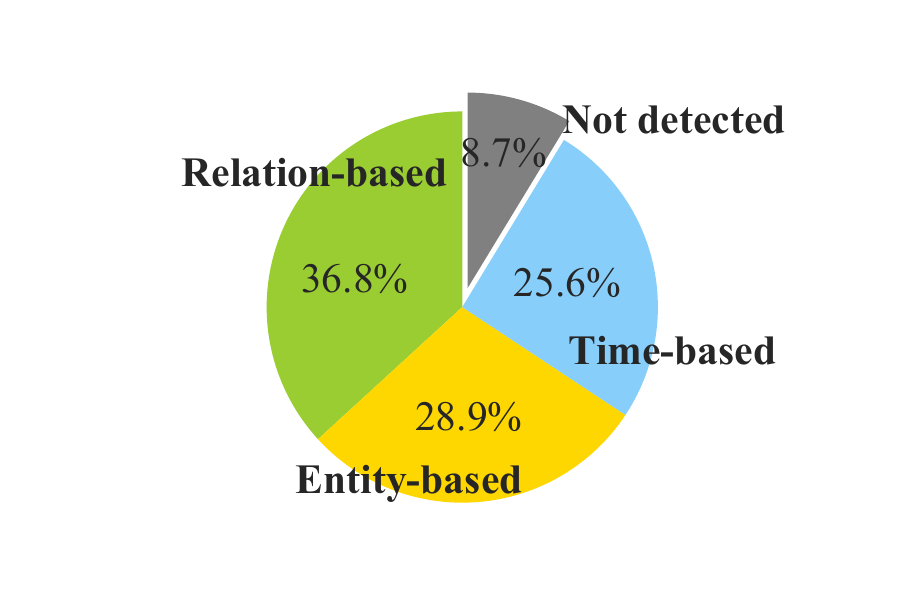}}
  \subfigure[Wikidata 12k]{\includegraphics[width=0.35\linewidth]{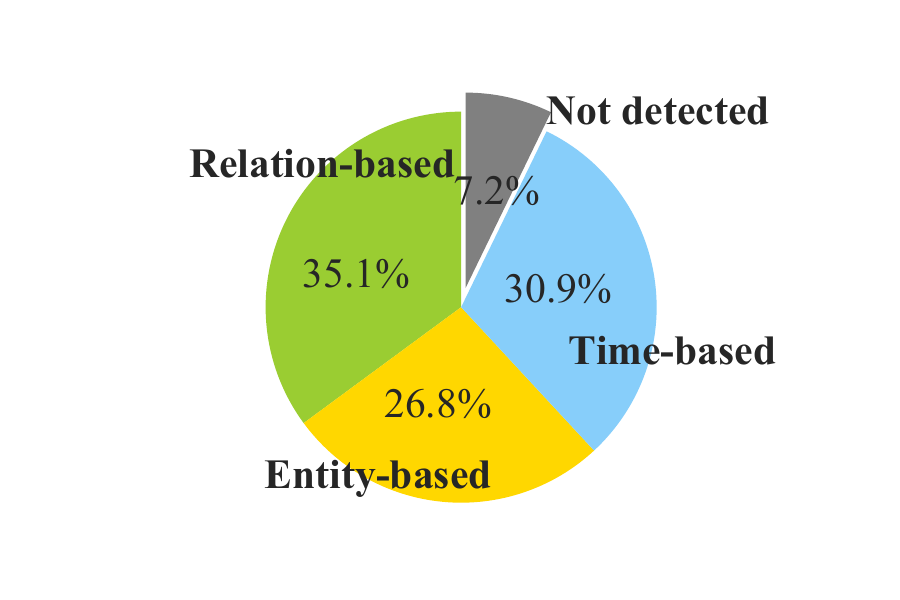}}
  \caption{The proportion of false negative samples detected by
the frequency-based filtering strategy in four real-world datasets.}
  \label{fig:selecting strategy}
\end{figure}

We only perform these filtering strategies for observed facts with
sparse local structures (i.e., $N(s) \leq k$ or $N(o) \leq k$) to
reduce the time complexity. To verify the effectiveness of our
strategies, we randomly remove 20\% facts from $\mathcal{G}$ and
calculate how many of them can be detected by our strategies. As
shown in Figure~\ref{fig:selecting strategy}, more than 90\% of
removed facts can be detected, highlighting the effectiveness of our
strategy in filtering potential false negatives.

Filtering out these potential false negatives can prevent the model
from being misled and enrich the sparse structure by further
identifying real false negatives. Moreover, since these strategies
have considered intrinsic patterns of TKG, the filtered samples
contain fine-grained information that enhances the model's ability
when used for fine-tuning.

\subsection{Hierarchical Scoring Algorithm}

Identifying real false negatives is challenging, for two reasons. 1)
TKGs have intricate semantic and temporal patterns brought by
diverse entity and relation categories, as well as time-evolving
topology. To accurately recognize real false negatives, it is
essential to account for these patterns adequately. 2) TKGs are
usually incomplete with noise graph structures, making the
identification inevitably have some mistakes. Thus, it is crucial to
assess the confidence of each fact to prevent the model from being
misled by low-confidence false negatives.

\begin{figure}[t]
  \centering
  \includegraphics[width=1\linewidth]{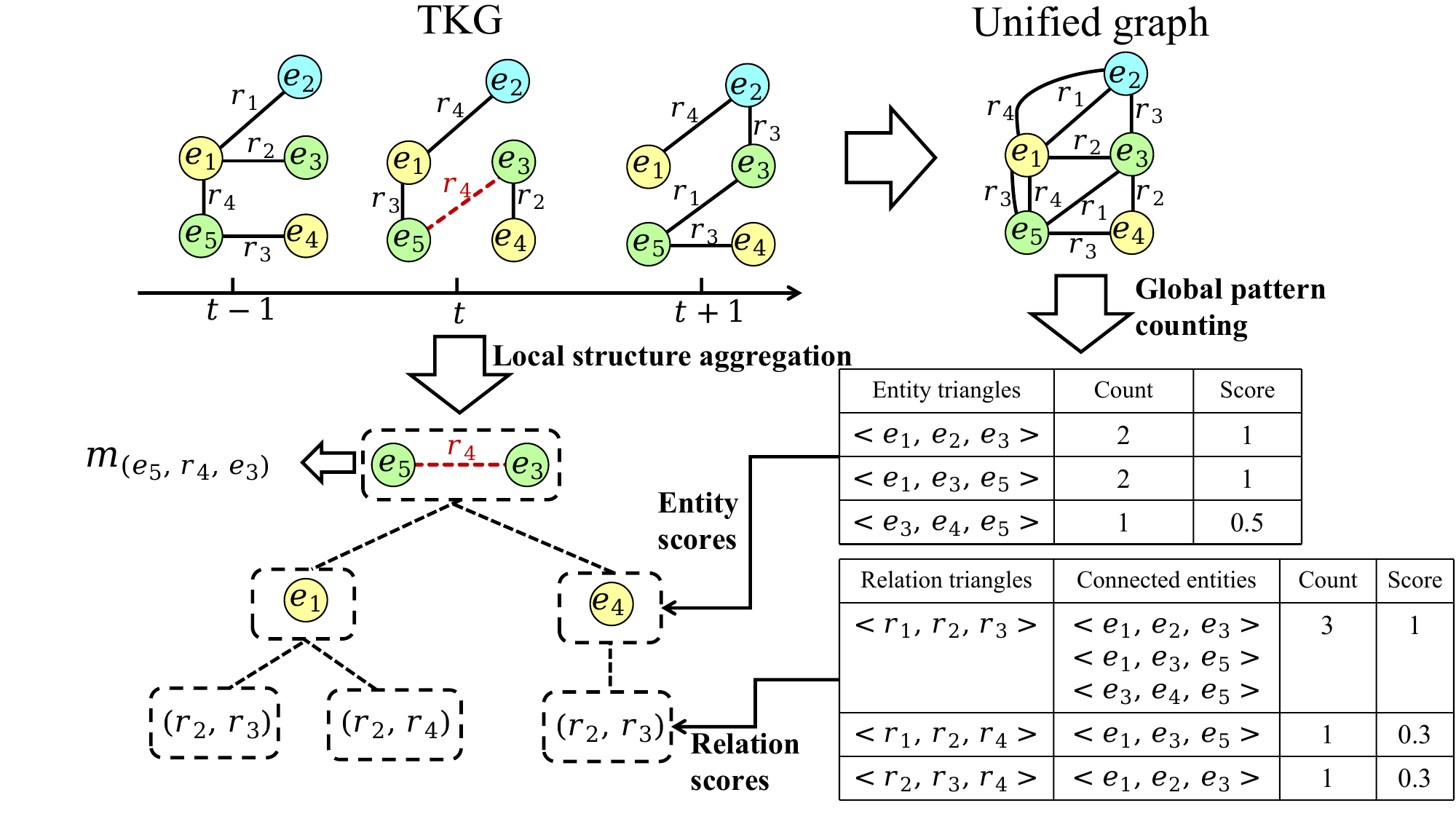}
  \caption{An example of the hierarchical scoring algorithm,
where the red dashed line denotes the potential false negative fact
that needs identification.}
  \label{fig:structured_pattern_scoring}
\end{figure}

In this part, we propose a novel hierarchical scoring algorithm,
assigning scores for potential false negatives to indicate their
possibility to be real false negatives. To incorporate both the
local structures and the global patterns of TKG while alleviating
the effect of skewed data distribution, our algorithm divides the
scoring process into two parts: global pattern counting and local
structure aggregation. As shown in
Figure~\ref{fig:structured_pattern_scoring}, the former generates
global scores via counting triangles on the unified graph, and the
latter aggregates the global score based on the local structure of a
sample. It uses a hierarchical structure (i.e., entity scores and
relation scores) to alleviate the complexity explosion brought by
combinations of node and edge categories. Additionally, a
perturbation technique is used to estimate the stabilization of the
scoring process, reducing the impact of noisy graph structures.

\textbf{Global pattern counting.} We represent the semantic patterns
of TKGs as triangle closures that include node and edge categories.
By identifying frequent triangles and estimating their intensity
(i.e., the likelihood that a third edge will complete the triangle
when two edges are present), we can effectively capture and quantify
these patterns. However, directly counting triangles in TKGs yields
a vast number of distinct triangles due to the mix of various
entities, relation types, and temporal orders, making the process
time-consuming and leading to sparse counts that poorly reflect
triangle intensity. To address this, we propose decomposing entities
and relations within triangles to reduce the complexity and using a
time-irrelevant graph to reduce noise from temporal ordering.

The time-irrelevant graph contains all the entity-relation-entity
combinations observed in TKG, represented as $\mathcal{G}^{'} =
\{(s, r, o) | \{(s,r,o,t) \in \mathcal{G} \}$. Since
$\mathcal{G}^{'}$ has been removed from the non-uniform distribution
of data in the time dimension, it can faithfully reflect the
preferences among entities and relations. To alleviate the
complexity explosion brought by the combination of entity and
relation types in triangle counting, we first anonymize the relation
types in $\mathcal{G}^{'}$ to count entity triangles. We define the
number of edges between entities as $N(e_1, e_2) = | \{(e_1,r,e_2) |
(e_1,r,e_2) \in \mathcal{G}^{'} \}|$ where $|\cdot|$ means the size
of the set. The number of triangles among entities can be
consequently defined as:
\begin{equation} \small
    C_e(e_1, e_2, e_3) = min(N(e_1, e_2) , N(e_2, e_3) , N(e_1, e_3)),
\end{equation}
which counts for the number of edges existing among three entities.
Since the anonymization of relation types highlights the
connectivity of the graph, $C_e$ can accurately reflect the
connection preference among entities, e.g., China and Japan have
more connections with South Korea than the Vatican. Finally, we
obtain the entity score by normalization as:
\begin{equation} \small
    S_e(e_1, e_2, e_3) = \frac{C_e(e_1, e_2, e_3) - min(C_e) + 1}{max(C_e) - min(C_e) + 1},
\end{equation}
where $max(C_e)$ and $min(C_e)$ are respectively the maximum number
and minimum number of entity triangles in $\mathcal{G}^{'}$.

The relation triangles indicate the interaction rules among entities
(e.g., the combination of relations ``launching an attack", ``call
for support", and ``impose sanctions" describes the hostile behavior
among three countries), and thus are important to identify the
missing valid facts. This motivates us to anonymize the entities in
$\mathcal{G}'$ to find relation triangles. Specifically, a relation
triangle consists of three relations that connect the entities
within an entity triangle. We define the count of each relation
triangle as how many different entity triangles are connected by it,
which is formally defined as:
\begin{equation} \small
    \begin{aligned}
    &C_r(r_1, r_2, r_3)\\ &=|\{ (e_i, e_j, e_k) | (e_i, r_1, e_j), (e_j, r_2, e_k), (e_i, r_3, e_k) \in \mathcal{G}'\}|.
    \end{aligned}
\end{equation}
The same normalization is then employed on $C_r$ to obtain the
relation score $S_r$. In summary, a high entity score suggests that
three entities are more likely to form a triangle, while a high
relation score indicates a stronger connection among the entities
through three specific relations. This effectively quantifies the
patterns present in the TKG.

\textbf{Local structure aggregation.} The validity of a fact depends
not only on its alignment with global semantic patterns but also on
its relevance to recent facts. While global scores summarize the
semantic patterns within TKGs, we propose aggregating them based on
the local structure of each fact to consider its short-term
relevance. As shown in Figure~\ref{fig:structured_pattern_scoring},
we reformulate the local structure of each potential false negative
fact $(s,r,o,t)$ into an entity layer and a relation layer, allowing
us to hierarchically aggregate the global scores. The entity layer
contains all entities that have interactions with both $s$ and $o$
within the time window $L_e$, denoted as $l_e(s,o,t)$. Therefore,
each entity $e$ in the entity layer can form a triangle with $s$ and
$o$, allowing us to estimate the validity of $(s,r,o,t)$ based on
intensities of entity triangles (i.e., the likelihood that $s-o$
will exist when $s-e$ and $o-e$ are present). To integrate relation
scores, for each entity $e \in l_e(s,o,t)$, we construct its
corresponding relation layer $l_r(s, o, t, e)$. Each item in
relation layer is represented as $(r_i, r_j)$, where $r_i$ are
relations existing between $s$ and $e$ within $L_e$, and $r_j$ are
relations existing between $o$ and $e$ within $L_e$. Therefore, each
item $(r_i, r_j)$ in the relation layer can form a triangle with
$r$, allowing us to estimate the validity of $(s,r,o,t)$ based on
intensities of relations triangles (i.e., the probability that $s$
and $o$ are connected by $r$ given that $s$ and $o$ have connected
with $e$ through $r_i$ and $r_j$). Afterward, we aggregate the
relation layer as:
\begin{equation} \small
    m_{e_i} = \sum_{(r_i, r_j) \in l_r(s, o, t, e_i)} \alpha_{i,j} \cdot S_r(r_i, r_j, r),
\end{equation}
where $e_i \in E(s,o,t)$ represents each node in the entity layer of
$(s,r,o,t)$. We use $\alpha$ as a time-aware weight to emphasize
facts that occurred more recently, which is defined as:
\begin{equation} \small
    \alpha_{i,j} = softmax(-|t_i - t_j|),
\end{equation}
where $t_i$ and $t_j$ are respectively the occurring time of facts
$(s,r_i,e)$ and $(e,r_j,o)$. We then aggregate the entity layer as:
\begin{equation} \small
    m_{(s,r,o,t)} = \sum_{e_i \in E(s,o,t)} m_{e_i} \cdot (1+S_e(s, o, e_i)),
\end{equation}
where $m_{(s,r,o,t)}$ is the confidence score of $(s,r,o,t)$.

\textbf{Score perturbation.} Each potential false negative fact
originates from an observed fact (see
Section~\ref{sec:frequency-based_filtering}), sharing a similar
local structure. Therefore, we can achieve the adaptive threshold by
comparing their scores. A potential false negative $f'$ will be
identified as a real false negative if $m_{f'}>m_{f}$ where $f$ is
its corresponding observed fact. However, the scoring process
inevitably has noise because of the incomplete graph structure. We
extend our algorithm with the perturbation technique and smooth
labels. For each potential false negative $f'$, we first slightly
perturb its corresponding hierarchical structures (e.g., randomly
repeat or remove items within layers or perturb the time-aware
weights), and then calculate a set of scores $M_{f'} = \{m^1_{f'},
m^2_{f'}, ..., m^k_{f'}\}$ based on these perturbed hierarchical
structures. If $mean(M_{f'})>m_{f}$, we will set $mean(M_{f'})$ as a
smooth label for $f'$, which emphasizes the samples with more robust
structures and more stable patterns. The remaining facts are
regarded as hard negative samples.

Notably, although data-driven methods can also be trained to
identify false negative facts, such as HyTE
\cite{DBLP:conf/emnlp/DasguptaRT18} and DE
\cite{DBLP:conf/aaai/GoelKBP20}, our algorithm has three main
advantages. First, our algorithm can generate reliable confidence
scores for these facts, providing more fine-grained information.
Second, our algorithm requires no additional training and thus is
more efficient. Third, our algorithm considers various patterns
within TKG, helping in achieving more accurate identification.
Although some data-driven methods also consider patterns, they
typically involve complex architectures that demand extensive
training time.

\subsection{Two-Stage Training}

The filtering strategies can identify potential false negatives,
while the hierarchical scoring algorithm further distinguishes
between real false negatives and hard negatives. On this basis, we
propose a two-stage training approach to shield the model from
imbalanced data and alleviate the model preferences.

\textbf{Pre-training.} Our filtering strategies have identified
potential false negatives that may be valid facts but are missing in
a TKG. To avoid misleading, we exclude these from being used as
negative samples during contrastive training. Formally, given an
untrained TKGC model $M^{(0)}$, we pre-train it using
\begin{equation} \small
    L_{p} = \sum_{(s,r,o,t) \in \mathcal{G}} -log(\frac{exp(p(s,r,o,t))}{\sum_{Neg(s,r,o,t)} exp(p(s,r,e,t))}) + \lambda,
\end{equation}
where each $Neg(s,r,o,t)$ are filtered negative facts specific to
$((s,r,o,t))$, defined as $Neg(s,r,o,t) = \{(s,r,e,t)| e \in
\mathcal{E}, (s,r,e,t) \notin EN_{(s,r,o,t)}\cup RN_{(s,r,o,t)}\cup
TN_{(s,r,o,t)}\}$. $\lambda$ is a regularization term and $p(\cdot)$
is the prediction score obtained by model $M^{(0)}$. The pre-trained
model is denoted as $M^{(1)}$.

Since the pre-training process avoids the misleading effects of
false negatives, the lower-ranked positive samples from the
pre-trained model $M^{(1)}$ accurately reveal which facts the model
is hard to learn (i.e., whose pattern it does not prefer), and thus
indicate model preferences. We select these low-ranked positive
samples as model-specific hard samples, emphasizing them during
training to correct misleading introduced by the model preferences.
Formally, for each positive sample $(s,r,o,t) \in \mathcal{G}$ with
its corresponding negative samples $(s,r,e,t) \in Neg(s,r,o,t)$, we
select $(s,r,o,t)$ as a model-specific hard sample if it is not
ranked higher than all of $(s,r,e,t)$ by $M^{(1)}$. We denote the
model-specific hard samples as $\mathcal{F}_m$ and use them for
fine-tuning.

\textbf{Fine-tuning.} We then fine-tune the model $M^{(1)}$ on the
identified real false negatives, hard negative samples, and the
model-specific hard samples to obtain the final model $M^{(2)}$.
This has three key purposes: 1) During fine-tuning, real false
negatives are treated as positive samples, helping to enrich the
data for entities and timestamps with sparse local structures. This
approach reduces the performance imbalance caused by uneven data
distribution. Meanwhile, the smooth labels generated through the
perturbation strategy emphasize facts with more robust structures
and more stable patterns, providing TKGC models with finer-grained
information. 2) Hard negative samples are facts that resemble
positive samples in pattern but are actually invalid, making them
difficult for the model to distinguish. Fine-tuning on these samples
enhances the model's discrimination ability. 3) Model-specific hard
samples reflect patterns that the model is hard to learn. Therefore,
fine-tuning on them can force the model to adapt to these
non-preferred patterns, enhancing the model's generalization ability
to various TKG patterns. Formally, we fine-tune the pre-trained
model $M^{(1)}$ using
\begin{equation} \small
    L_f = \sum_{f \in \mathcal{G}^p} -l_f log \sigma(p(f)) + \sum_{f' \in \mathcal{G}^n} - log \sigma(p(f')),
\end{equation}
where $\mathcal{G}^p$ contains real false negatives and
model-specific hard samples, while $\mathcal{G}^n$ contains hard
negative samples. $\sigma(\cdot)$ is the sigmoid function and $l_f =
mean(M_f)$ is the smooth label of fact $f$. The fine-tuned model
$M^{(2)}$ thus has eliminated the effects of imbalanced data and
model preferences.

\subsection{Complexity Analysis}

The additional time complexity introduced by \textit{Booster} comes
from the filtering strategy and the hierarchical scoring algorithm.
During filtering potential false negatives, for each observed fact,
it takes $O(n_r)$ time to achieve the relation-based filtering,
where $n_r$ is the number of relations that co-occur with relation
$r$. The time for entity-based filtering is $O(n_{s,o})$ where
$n_{s,o}$ is the number of entities that have interacted with
entities $s$ and $o$, and the time for time-based filtering is
$O(L_t)$. For the hierarchical scoring, by iterating over the
unified graph, the time for calculating the entity score and
relation score is $O(|\mathcal{E}|^{2})$ where $|\mathcal{E}|$ is
the number of entities within TKG. During local structure
aggregation, it takes $O(n_{s,o} \cdot q_{s,e}\cdot q_{o,e})$ time
for each sample where $q_{s,e}$ is the number of relations between
$s$ and the co-interacted entity $e$.

\section{Experiments}

We conduct experiments on five benchmark datasets to answer the
following research questions: \textbf{RQ1}: Can \textit{Booster}
improve the performance of existing models? \textbf{RQ2}: How does
each component of \textit{Booster} contribute to performance
improvement? \textbf{RQ3}: Is \textit{Booster} efficient?
\textbf{RQ4}: Can \textit{Booster} improves the balance and
stabilization of the performance?

\begin{table}[t] \footnotesize
\caption{Statistics of datasets.} \centering
  \begin{tabular}{l||cccc}
    \hline
    \textbf{Dataset} & $\mathcal{|E|}$ & $\mathcal{|R|}$ & $\mathcal{|T|}$ & $|\mathcal{F}|$\\
    \hline
    \textbf{ICEWS 14} &7,128 &230 &365 &90,730 \\
    \textbf{ICEWS 05-15} &10,488 &251 &4,017 &461,329 \\
    \textbf{YAGO 11k} &10,623 &10 &2,801 &20,507 \\
    \textbf{Wikidata 12k} &12,554 &24 &2,270 &40,621 \\
    \textbf{GDELT} &500 &20 &366 &3,419,607 \\
    \hline
  \end{tabular}
  \label{tab:datasets}
\end{table}

\textbf{Datasets.} We evaluate\textit{Booster} on five benchmark TKG
datasets, which are from ICEWS \cite{ICEWS}, YAGO
\cite{DBLP:conf/www/SuchanekKW07}, Wikidata
\cite{DBLP:conf/semweb/ErxlebenGKMV14}, and GDELT
\cite{DBLP:conf/isa/LeetaruS13}. ICEWS contains interactions among
political people and countries with time annotations. We use two
subsets of it, i.e., ICEWS 14 and ICEWS 05-15, which contain
knowledge in 2014 and knowledge from 2005 to 2015 respectively. YAGO
is a knowledge base that contains common sense knowledge. YAGO 11k
is formed by selecting knowledge that contains the top-10 frequent
time-sensitive relations from YAGO. Wikidata is an open knowledge
base driven from Wikipedia and Wikidata 12k is a subset of it. GDELT
is a large political knowledge base. We split each dataset as
train/validation/test set with the proportion of 8:1:1. The detailed
statistics of these datasets are shown in Table~\ref{tab:datasets}.

\textbf{Comparison models.} We compare \textit{Booster} with
existing data augmentation strategies for temporal graphs (MeTA
\cite{DBLP:conf/nips/WangCLDWBH21}) and knowledge graphs (DeMix
\cite{DBLP:conf/semweb/ChenZYCT23}, NSCaching
\cite{DBLP:journals/vldb/ZhangY021}, and KG-Mixup
\cite{DBLP:conf/www/ShomerJ0T23}). We use five popular TKGC models
as backbones to evaluate the effectiveness of these data
augmentation strategies: HyTE \cite{DBLP:conf/emnlp/DasguptaRT18},
TA \cite{DBLP:conf/emnlp/Garcia-DuranDN18}, DE
\cite{DBLP:conf/aaai/GoelKBP20}, TNT
\cite{DBLP:conf/iclr/LacroixOU20}, and TEMP
\cite{DBLP:conf/emnlp/WuCCH20}.

\begin{table*}[t]
\caption{Performance comparison of baseline models. The best results
are boldfaced and ``DA" means data augmentation.} \centering
\scalebox{0.75}{
\begin{tabular}{c|c|ccc|ccc|ccc|ccc}
\hline &\textbf{Dataset} &\multicolumn{3}{c|}{\textbf{ICEWS 14}}
&\multicolumn{3}{c|}{\textbf{ICEWS 05-15}}
&\multicolumn{3}{c|}{\textbf{YAGO 11k}}
&\multicolumn{3}{c}{\textbf{Wikidata 12k}}\\
\hline \textbf{TKGC models} &\textbf{DA strategies} &\emph{MRR}
&\emph{Hits@1}   &\emph{Hits@3} &\emph{MRR} &\emph{Hits@1}
&\emph{Hits@3} &\emph{MRR} &\emph{Hits@1}   &\emph{Hits@3}
&\emph{MRR} &\emph{Hits@1}   &\emph{Hits@3} \\
\hline

&Without DA &0.297 &0.108 &0.416 &0.316 &0.116 &0.445 &0.134 &0.032 &0.181 &0.191 &0.107 &0.208\\
&MeTA &0.293 &0.105 &0.410 &0.319 &0.117 &0.449 &0.132 &0.031 &0.178 &0.186 &0.105 &0.201\\
&DeMix &0.301 &0.113 &0.412 &0.323 &0.126 &0.447 &0.136 &0.035 &0.182 &0.193 &0.108 &0.211\\
HyTE &NSCaching &0.295 &0.107 &0.414 &0.320 &0.124 &0.445 &0.130 &0.029 &0.179 &0.188 &0.105 &0.206\\
&KG-Mixup &0.308 &0.133 &\textbf{0.420} &0.325 &0.145 &0.446 &0.136 &0.037 &0.175 &0.192 &0.107 &0.210\\
&\textit{Booster} &\textbf{0.323} &\textbf{0.176} &0.417 &\textbf{0.341} &\textbf{0.223} &\textbf{0.450} &\textbf{0.142} &\textbf{0.051} &\textbf{0.182} &\textbf{0.199} &\textbf{0.112} &\textbf{0.219} \\
\cline{3-14} & &\multicolumn{1}{c}{Improve: 8.7\%}
&\multicolumn{2}{c|}{P-value: 0.0114} &\multicolumn{1}{c}{Improve:
7.9\%} &\multicolumn{2}{c|}{P-value: 0.0101}
&\multicolumn{1}{c}{Improve: 5.9\%} &\multicolumn{2}{c|}{P-value:
0.0225}
&\multicolumn{1}{c}{Improve: 4.1\%} &\multicolumn{2}{c}{P-value: 0.0217} \\
\hline

&Without DA &0.501 &0.392 &0.569 &0.484 &0.366 &0.546 &0.119 &0.084 &0.117 &0.212 &0.123 &0.242\\
&MeTA &0.496 &0.388 &0.565 &0.486 &0.367 &0.544 &0.112 &0.078 &0.113 &0.209 &0.119 &0.246\\
&DeMix &0.508 &0.397 &0.573 &0.491 &0.370 &0.545 &0.116 &0.081 &0.118 &0.214 &0.124 &0.246\\
DE &NSCaching &0.503 &0.394 &0.570 &0.482 &0.365 &0.541 &0.115 &0.080 &0.116 &0.217 &0.125 &0.248\\
&KG-Mixup &0.507 &0.398 &0.565 &0.492 &0.372 &0.545 &0.117 &0.082 &0.115 &0.215 &0.122 &0.244\\
&\textit{Booster} &\textbf{0.521} &\textbf{0.410} &\textbf{0.578} &\textbf{0.510} &\textbf{0.388} &\textbf{0.548} &\textbf{0.124} &\textbf{0.086} &\textbf{0.118} &\textbf{0.221} &\textbf{0.126} &\textbf{0.252} \\
\cline{3-14} & &\multicolumn{1}{c}{Improve: 3.9\%}
&\multicolumn{2}{c|}{P-value: 0.0170} &\multicolumn{1}{c}{Improve:
3.5\%} &\multicolumn{2}{c|}{P-value: 0.0217}
&\multicolumn{1}{c}{Improve: 4.2\%} &\multicolumn{2}{c|}{P-value:
0.0233}
&\multicolumn{1}{c}{Improve: 4.2\%} &\multicolumn{2}{c}{P-value: 0.0213} \\

\hline

&Without DA &0.409 &0.295 &0.466 &0.492 &0.376 &0.544 &0.110 &0.072 &0.108 &0.188 &0.109 &0.210\\
&MeTA &0.405 &0.293 &0.460 &0.493 &0.377 &0.542 &0.112 &0.078 &0.104 &0.186 &0.107 &0.208\\
&DeMix &0.411 &0.292 &0.473 &0.498 &0.380 &0.545 &0.106 &0.071 &0.103 &0.184 &0.106 &0.210\\
TA &NSCaching &0.401 &0.289 &0.460 &0.493 &0.375 &0.550 &0.110 &0.077 &0.112 &0.180 &0.103 &0.211\\
&KG-Mixup &0.412 &0.290 &0.475 &0.502 &0.381 &0.544 &0.113 &0.075 &0.110 &0.191 &0.110 &0.210\\
&\textit{Booster} &\textbf{0.421} &\textbf{0.298} &\textbf{0.484} &\textbf{0.513} &\textbf{0.387} &\textbf{0.566} &\textbf{0.123} &0.081 &\textbf{0.115} &\textbf{0.205} &\textbf{0.113} &\textbf{0.214}\\
\cline{3-14} & &\multicolumn{1}{c}{Improve: 2.9\%}
&\multicolumn{2}{c|}{P-value: 0.0156} &\multicolumn{1}{c}{Improve:
4.2\%} &\multicolumn{2}{c|}{P-value: 0.0139}
&\multicolumn{1}{c}{Improve: 6.9\%} &\multicolumn{2}{c|}{P-value:
0.0218}
&\multicolumn{1}{c}{Improve: 3.5\%} &\multicolumn{2}{c}{P-value: 0.0185} \\

\hline

&Without DA &0.601 &0.478 &0.681 &0.680 &0.553 &0.769 &0.186 &0.126 &0.189 &0.330 &0.227 &0.359\\
&MeTA &0.602 &0.479 &0.682 &0.676 &0.548 &0.766 &0.184 &0.125 &0.184 &0.327 &0.225 &0.358\\
&DeMix &0.606 &0.479 &0.684 &0.682 &0.555 &0.769 &0.188 &0.124 &0.189 &0.331 &0.229 &0.360\\
TEMP &NSCaching &0.598 &0.476 &0.677 &0.678 &0.550 &0.764 &0.180 &0.121 &0.187 &0.325 &0.220 &0.356\\
&KG-Mixup &0.603 &0.480 &0.676 &0.684 &0.552 &0.770 &0.186 &0.124 &0.190 &0.332 &0.228 &0.360\\
&\textit{Booster} &\textbf{0.623} &\textbf{0.485} &\textbf{0.690} &\textbf{0.697} &\textbf{0.559} &\textbf{0.779} &\textbf{0.194} &\textbf{0.130} &\textbf{0.195} &\textbf{0.340} &\textbf{0.237} &\textbf{0.362}\\
\cline{3-14} & &\multicolumn{1}{c}{Improve: 3.6\%}
&\multicolumn{2}{c|}{P-value: 0.0151} &\multicolumn{1}{c}{Improve:
2.5\%} &\multicolumn{2}{c|}{P-value: 0.0219}
&\multicolumn{1}{c}{Improve: 4.3\%} &\multicolumn{2}{c|}{P-value:
0.0197}
&\multicolumn{1}{c}{Improve: 3.3\%} &\multicolumn{2}{c}{P-value: 0.0189} \\
\hline

&Without DA &0.614 &0.532 &0.656 &0.658 &0.588 &0.712 &0.185 &0.127 &0.183 &0.331 &0.233 &0.357\\
&MeTA &0.608 &0.529 &0.649 &0.650 &0.582 &0.706 &0.183 &0.127 &0.175 &0.332 &0.234 &0.355\\
&DeMix &0.615 &0.530 &0.660 &0.661 &0.590 &0.714 &0.188 &0.129 &0.185 &0.331 &0.231 &0.360\\
TNT &NSCaching &0.605 &0.526 &0.649 &0.652 &0.584 &0.709 &0.180 &0.123 &0.176 &0.327 &0.228 &0.355\\
&KG-Mixup &0.619 &0.537 &0.661 &0.663 &0.591 &0.710 &0.187 &0.126 &0.188 &0.335 &0.234 &0.359\\
&\textit{Booster}
&\textbf{0.636} &\textbf{0.557} &\textbf{0.678} &\textbf{0.679} &\textbf{0.602} &\textbf{0.728} &\textbf{0.195} &\textbf{0.131} &\textbf{0.201} &\textbf{0.342} &\textbf{0.239} &\textbf{0.367} \\
\cline{3-14} & &\multicolumn{1}{c}{Improve: 3.5\%}
&\multicolumn{2}{c|}{P-value: 0.0145} &\multicolumn{1}{c}{Improve:
3.2\%} &\multicolumn{2}{c|}{P-value: 0.0123}
&\multicolumn{1}{c}{Improve: 5.4\%} &\multicolumn{2}{c|}{P-value:
0.0119}
&\multicolumn{1}{c}{Improve: 3.3\%} &\multicolumn{2}{c}{P-value: 0.0168} \\

\hline
\end{tabular}}
\label{tab:TKGC}
\end{table*}

\textbf{Implementation details.} We use the released official
implementation of existing TKGC models as the backbone. For each
model, we tune its hyper-parameters using a grid search, where the
best hyper-parameter settings are selected with the best MRR on the
validation set. We create 100 mini-batches for each epoch during
training, and the number of epochs is set as 1000. The models are
pre-trained in the first 20 epochs and then fine-tuned in the rest
epochs with early stopping. The learning rate is set as 0.001. For
all models, we set the representation dimension $d$ as 200, the size
of negative sampling as 50, and the data pre-processing is unified
as in TNT \cite{DBLP:conf/iclr/LacroixOU20} to achieve fair
comparison. The time windows $L_r$, $L_e$, and $L_t$ are selected
from $\{1, 3, 5, 10, 20\}$. We use Adagrad
\cite{DBLP:journals/ijics/LydiaF19} for optimization and all
experiments are conducted on a 64-bit machine with Nvidia TITAN RTX.
Besides MRR, we also use Hits@k as the metric which is defined as
$Hits@k = \frac{1}{|Test|} \sum_{(s,r,o,t) \in Test}
{ind(rank(s,r,o,t) \leq k)}$, where $ind()$ is 1 if the inequality
holds and 0 otherwise. Our source code is available at
\url{https://github.com/zjs123/Booster}.

\begin{figure*}[t]
\centering \scalebox{0.88}{
\subfigure[]{\includegraphics[width=0.29\linewidth]{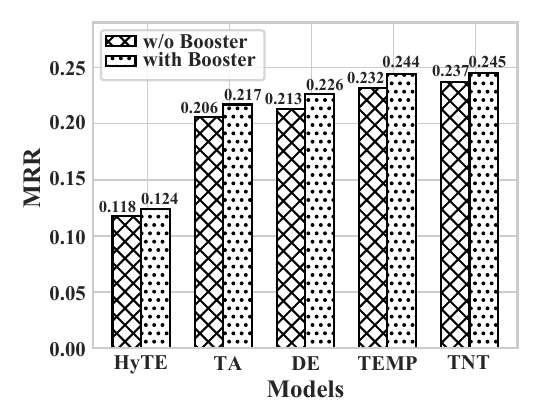}}\quad
\subfigure[]{\includegraphics[width=0.27\linewidth]{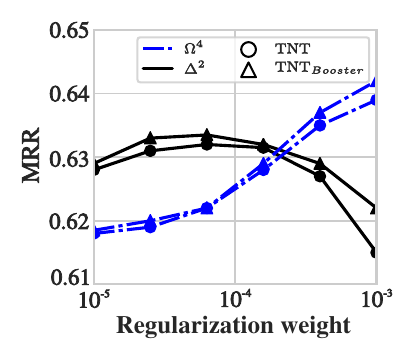}}\quad
\subfigure[]{\includegraphics[width=0.27\linewidth]{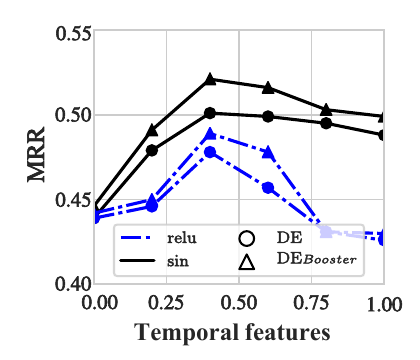}}}
\caption{(a) Performance of different methods on the GDELT dataset.
(b) Performance improvements of \textit{Booster} with varying
hyper-parameters of the TNT model on the ICEWS 14 dataset. (c)
Performance improvements of \textit{Booster} with varying
hyper-parameters of the DE model on the ICEWS 14 dataset.}
\label{fig:GDELT_TNT_DE_param}
\end{figure*}

\begin{figure*}[t]
\centering \scalebox{0.88}{
\subfigure[]{\includegraphics[width=0.27\linewidth]{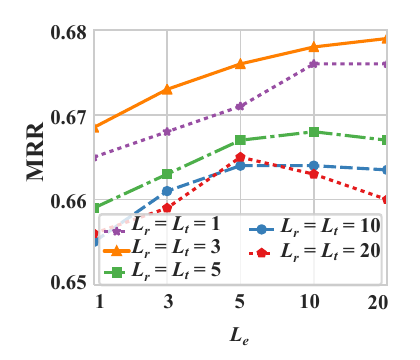}}\qquad
\subfigure[]{\includegraphics[width=0.27\linewidth]{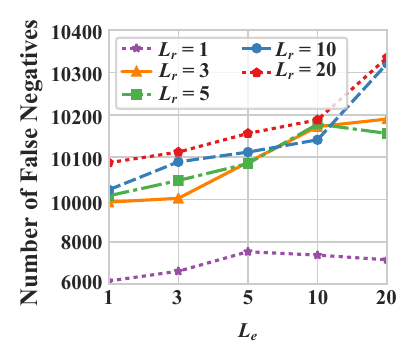}}\quad
\subfigure[]{\includegraphics[width=0.27\linewidth]{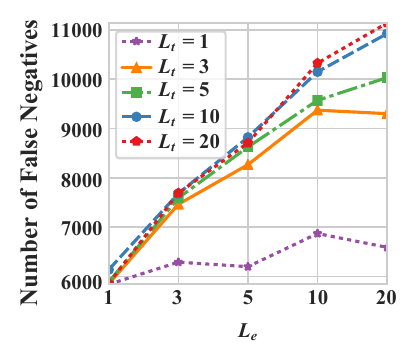}}}
\caption{(a) Performance of TNT$_\textit{Booster}$ with varying
hyper-parameters on the ICEWS 05-15 dataset. (b) The number of
identified false negatives with different $L_e$ and $L_r$. (c) The
number of identified false negatives with different $L_e$ and
$L_t$.} \label{fig:param_analysis}
\end{figure*}

\subsection{Overall Evaluation (RQ1)}
\label{sec:RQ1}

\textbf{Accuracy.} Table~\ref{tab:TKGC} shows the performance of
existing TKGC models with different data augmentation strategies. We
can see that: (1) \textit{Booster} can bring performance
improvements for all backbone models. The maximum improvement is
8.7\% and the average improvement is 4.5\% with statistical
significance, demonstrating that \textit{Booster} can seamlessly
adapt to existing models to improve the quality of the learned
representations. The HyTE model gets the largest improvement (8.7\%
in MRR on the ICEWS 14 dataset and 7.9\% in MRR on the ICEWS 05-15
dataset). This is because HyTE independently learns representations
in different timestamps and thus is more sensitive to the imbalanced
data distribution. \textit{Booster} can enrich the sparse structure
by identifying false negatives and thus achieve improvement. (2)
Compared with existing graph data augmentation strategies,
\textit{Booster} can achieve higher performance improvement for most
backbone models. Its improvement outperforms the powerful knowledge
graph data augmentation strategies such as KG-Mixup and NSCaching,
highlighting its effectiveness. One interesting observation is that
some previous data augmentation strategies may even result in
performance degradation such as MeTA and NSCaching. This is because
they fail to fully consider TKGs' complex semantic patterns and
temporal relevance when filtering false negatives and generating new
samples, and thus bring noise to the TKGC models. (3)
\textit{Booster} shows performance improvements on all of the
datasets, which demonstrates that it can adaptively handle
heterogeneous temporal knowledge from different fields. We notice
that the performance gains for a model may be different across
datasets. The average improvement of different models on YAGO 11k is
3.9\%, while on Wikidata 12k is 2.4\%. This is because the sparsity
of datasets can affect the performance gains. As shown in
Figure~\ref{fig:GDELT_TNT_DE_param}(a), results on the GDELT dataset
verifies \textit{Booster} can also achieve performance improvement
on large graphs.

\textbf{Hyper-parameter sensitivity.} In
Figure~\ref{fig:GDELT_TNT_DE_param}(b) and (c), $\Omega^4$ and
$\Delta^2$ are hyper-parameters of TNT and DE. $sin$ and $relu$ are
two nonlinear functions, and temporal features mean the percentage
of time-sensitive representations. We can see that \textit{Booster}
can achieve improvement with different hyper-parameters of the
original model. Figure~\ref{fig:param_analysis}(a) shows how the
hyper-parameters of \textit{Booster} affect its effectiveness. We
can see that when $L_e$ increases, $MRR$ keeps increasing at first
because the larger $L_e$ helps to achieve more accurate scoring of
facts. However, $MRR$ drops when $L_t$ and $L_r$ are large. This is
because the large $L_t$ and $L_r$ will extensively increase the
number of filtered potential false negatives and thus bring more
noise. Finally, Figure~\ref{fig:param_analysis}(b) and (c) show how
the number of identified real false negatives changes with
hyper-parameters. We can see that when $L_r$ and $Lt$ are large, the
number of identified real false negatives increases significantly,
but when $L_r = Lt = 1$, its number gradually decreases when $L_e$
increases, which meets our conjecture that large $L_t$ and $L_r$
will bring more noise.

\subsection{Effect of Each Component (RQ2)}
\label{sec:RQ2}

\begin{table}[t] \footnotesize
\centering \caption{Results of ablation study.} \scalebox{0.85}{
\begin{tabular}{c|cc|cc}
\hline \multicolumn{1}{c|}{\textbf{Dataset}}         &
\multicolumn{2}{c|}{\textbf{ICEWS 14}}                             &
\multicolumn{2}{c}{\textbf{Wikidata 12k}}
\\ \hline \multicolumn{1}{c|}{{\textbf{Variants}}}
 & \emph{Hit@1}          & \emph{Hit@10}
 & \emph{Hit@1}          & \emph{Hit@10}\\
\hline w/o identified false negatives
&0.545 &0.774 &0.334 &0.541 \\
\cline{1-1} w/o identified hard negatives
&0.544 &0.772 &0.332 &0.538 \\
\cline{1-1} w/o model-specific hard samples
&0.544 &0.772 &0.332 &0.538 \\
\cline{1-1} w/o entity scores
&0.548 &0.775 &0.330 &0.539 \\
\cline{1-1} w/o relation scores
&0.543 &0.771 &0.329 &0.537 \\
\cline{1-1} w/o smooth labels
&0.533 &0.768 &0.331 &0.537 \\
\cline{1-1} w/o smooth labels + entity scores
&0.523 &0.760 &0.324 &0.534 \\
\cline{1-1} w/o smooth labels + relation scores
&0.521 &0.757 &0.322 &0.530 \\
\hline \cline{1-1} TNT$_\textit{Booster}$
&\textbf{0.557} &\textbf{0.781} &\textbf{0.342} &\textbf{0.547} \\
\hline
\end{tabular}}
\label{tab:ablation}
\end{table}

\begin{figure*}[t]
\centering \scalebox{0.95}{
\subfigure[]{\includegraphics[width=0.24\linewidth]{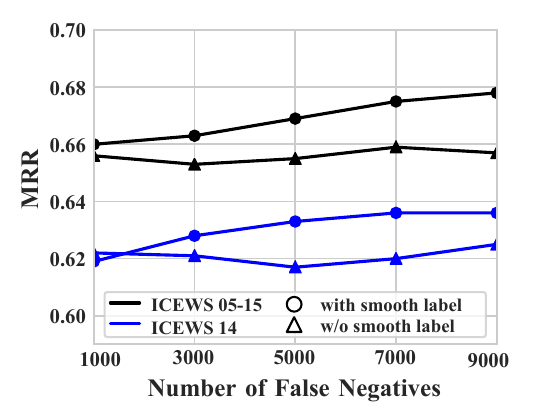}}
\subfigure[]{\includegraphics[width=0.24\linewidth]{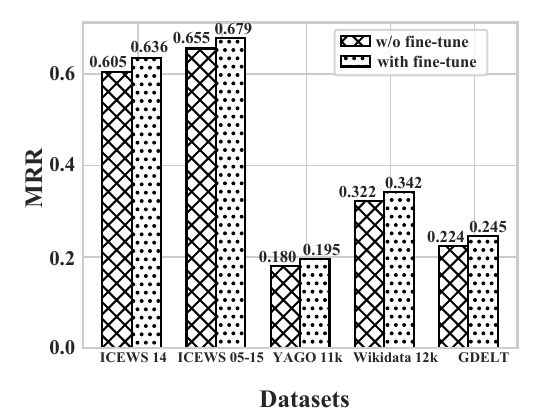}}
\subfigure[]{\includegraphics[width=0.22\linewidth]{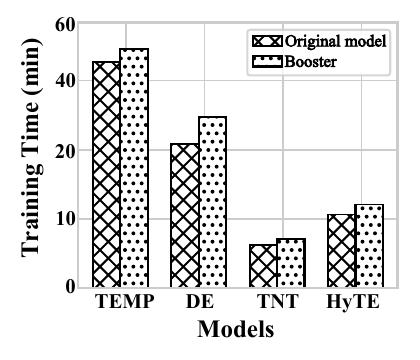}}\quad
\subfigure[]{\includegraphics[width=0.22\linewidth]{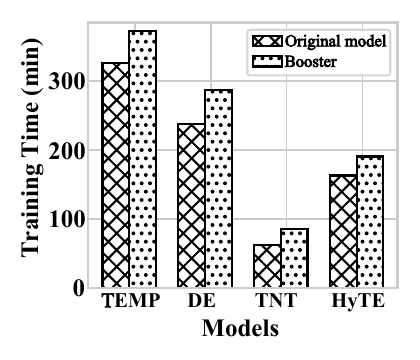}}}
\caption{(a) The sensitivity of model performance to the identified
false negatives. (b) Performance of the TNT model with and without
fine-tuning. (c) The training time of different models on the ICEWS
14 dataset. (d) The training time of different models on the ICEWS
05-15 dataset.} \label{fig:smooth_tune_traing_time}
\end{figure*}

\textbf{Ablation study.} Table~\ref{tab:ablation} shows the ablation
study results of \textit{Booster}. we can see that including
identified false negatives, hard negatives, and model-specific
challenging samples during fine-tuning all contribute to
performance. Removing the smooth label may cause performance
degradation since the false negatives with low confidence will
mislead the model during fine-tuning. Only removing entity scores or
relation scores leads to small degradation, which also demonstrates
the effectiveness of the smooth label in handling noises brought by
false negative identification. However, when the smooth label is
removed, removing scores will largely damage the performance, which
shows their benefit for accurate identification. We further show the
sensitivity of model performance to the identified false negatives
in Figure~\ref{fig:smooth_tune_traing_time}(a), where the variant
without smooth label gets worse and more fluctuated performance. In
Figure~\ref{fig:smooth_tune_traing_time}(b), we can see that
fine-tuning the model is necessary for performance improvement. This
has two reasons: 1) The pre-training is only based on a partial
structure of the original graph, so it cannot accurately express the
semantics of TKGs. 2) The fine-tuning samples are refined with
various pattern-aware heuristics and thus enhance the model's
generalization ability to various TKG patterns. Moreover, the
identified false negatives can enrich the sparse graph structure,
and thus improve the performance for long-tail entities and
timestamps.

\begin{figure*}[t]
\centering \scalebox{0.88}{
\subfigure[]{\includegraphics[width=0.27\linewidth]{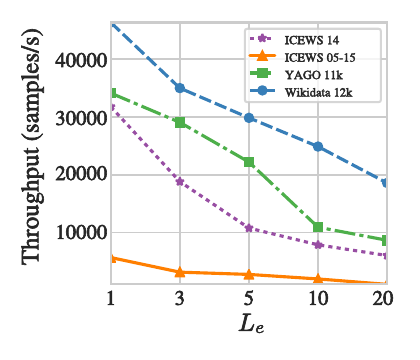}}
\subfigure[]{\includegraphics[width=0.27\linewidth]{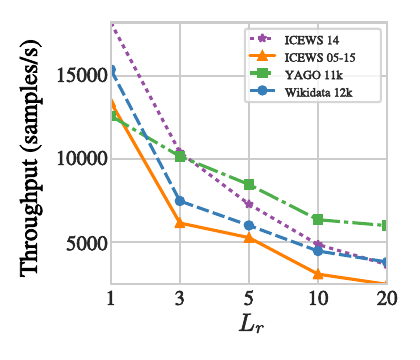}}
\subfigure[]{\includegraphics[width=0.27\linewidth]{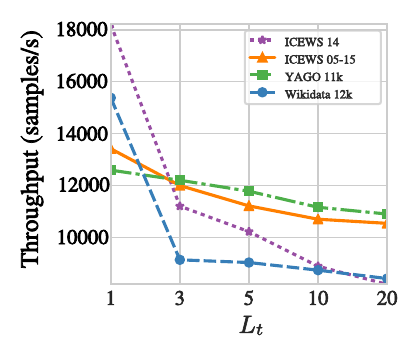}}}
\caption{(a) Throughput of the \texttt{Booster} framework for
processing different datasets with varying $L_e$. (b) Throughput of
the \texttt{Booster} framework for processing different datasets
with varying $L_r$.(c) Throughput of the \texttt{Booster} framework
for processing different datasets with varying $L_t$.}
\label{fig:efficency}
\end{figure*}

\begin{table}[t] \footnotesize
\centering \caption{Performance comparison of variants.}
\label{tab:compare} \scalebox{0.85}{
\begin{tabular}{c|cc|cc}
\hline \multicolumn{1}{c|}{\textbf{Dataset}}         &
\multicolumn{2}{c|}{\textbf{ICEWS 05-15}} &
\multicolumn{2}{c}{\textbf{Wikidata 12k}}
\\ \hline \multicolumn{1}{c|}{{\textbf{Variants}}}
 & \emph{Hit@1}          & \emph{Hit@10}
 & \emph{Hit@1}          & \emph{Hit@10}\\
\hline Identifying with DE
&0.595 &0.814 &0.234 &\textbf{0.548} \\
\cline{1-1} Identifying with TEMP
&0.596 &0.816 &0.235 &0.545 \\
\cline{1-1} Self-training
&0.582 &0.791 &0.230 &0.539 \\
\cline{1-1} Neighbor filtering
&0.587 &0.797 &0.230 &0.540 \\
\cline{1-1} Recent active filtering
&0.581 &0.790 &0.228 &0.538 \\
\hline \cline{1-1} TNT$_\textit{Booster}$
&\textbf{0.602} &\textbf{0.823} &\textbf{0.239} &0.547 \\
\hline
\end{tabular}
} \label{tab:variants}
\end{table}

\textbf{Comparison with variants.} Table~\ref{tab:variants} shows
the performance of the variants of \textit{Booster}. To verify the
effectiveness of the hierarchical scoring algorithm, we replace it
with the pre-trained DE and TEMP models to identify real false
negatives. We can see that they fail to outperform our original
framework. Moreover, as shown in
Figure~\ref{fig:smooth_tune_traing_time}(c) and (d), retraining
these models is time-consuming. The observations show the
superiority of our proposed scoring algorithm. Furthermore, we
directly employ self-training on the TNT model. We can see that the
performance degrades significantly, which shows the necessity of
considering model preferences. Finally, to verify the effectiveness
of the filtering strategies, we compare our strategy with widely
used false negative filtering strategies. Either selecting neighbor
entities with sparse local structures or selecting the most recent
active entities fail to achieve competitive performance to
TNT$_\textit{Booster}$.

\begin{figure*}[t]
\centering \scalebox{0.95}{
\subfigure[]{\includegraphics[width=0.24\linewidth]{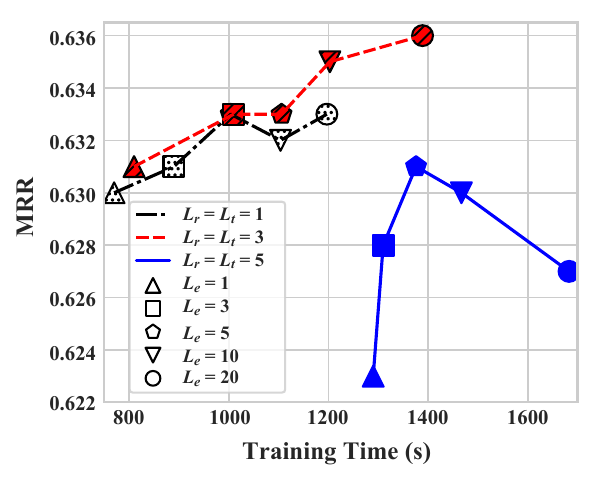}}
\subfigure[]{\includegraphics[width=0.24\linewidth]{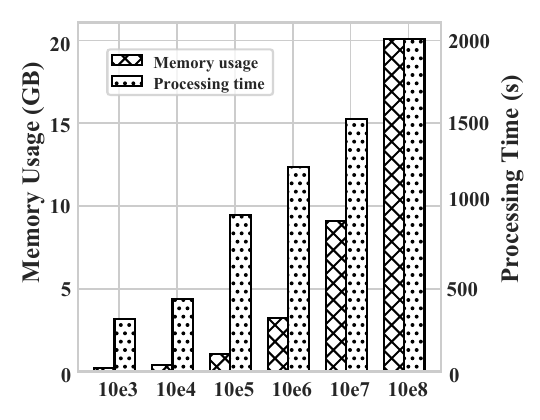}}
\subfigure[]{\includegraphics[width=0.24\linewidth]{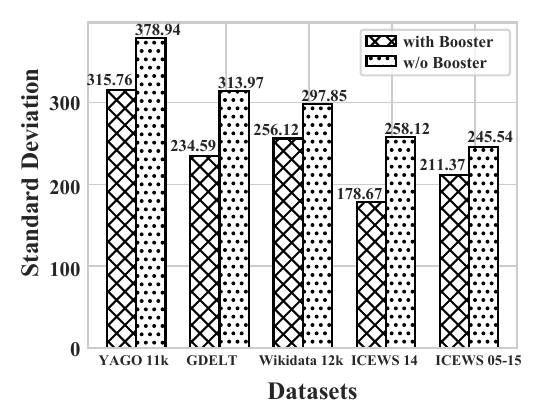}}
\subfigure[]{\includegraphics[width=0.24\linewidth]{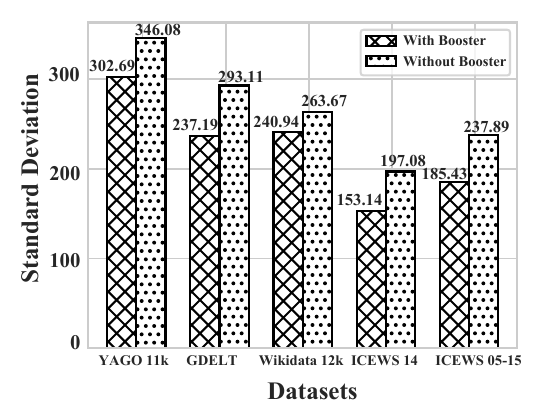}}}
\caption{(a) MRR w.r.t. training time on the ICEWS 14 dataset. (b)
Scalability of our proposed algorithm. (c) MRR variance on
timestamps obtained by the TEMP model. (d) MRR variance on
timestamps obtained by the TNT model.} \label{fig:MRR_scal_STD_Time}
\end{figure*}

\begin{figure*}[t]
\centering \scalebox{0.95}{
\subfigure[]{\includegraphics[width=0.24\linewidth]{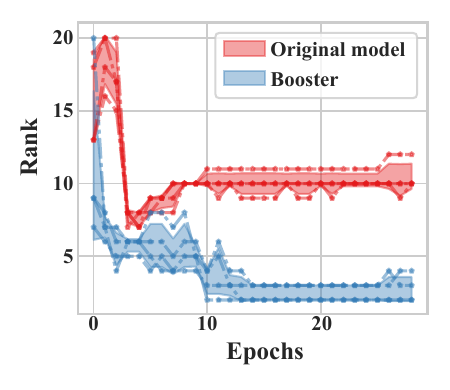}}
\subfigure[]{\includegraphics[width=0.24\linewidth]{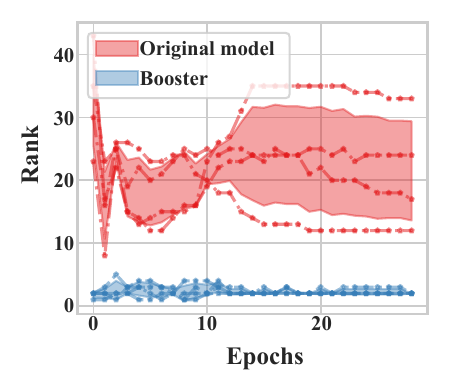}}
\subfigure[]{\includegraphics[width=0.24\linewidth]{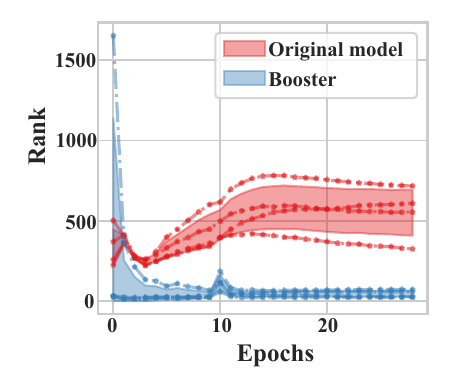}}
\subfigure[]{\includegraphics[width=0.24\linewidth]{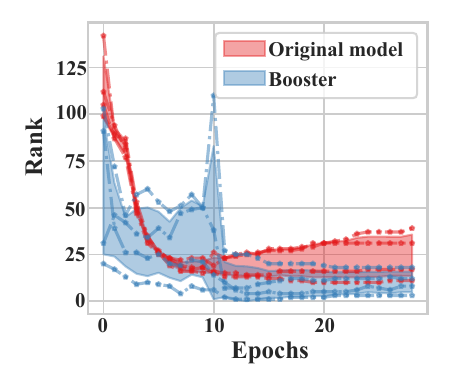}}}
\caption{Change of $rank$ of four test samples that are hard to be
optimized by the TEMP model.} \label{fig:case_example}
\end{figure*}

\subsection{Efficiency (RQ3)}

\textbf{Comparison with baselines.}
Figure~\ref{fig:smooth_tune_traing_time}(c) and (d) show the time
consumption of baseline models when trained with the
\textit{Booster} framework. We can see that the training time is
only increased by 1/10 and 1/5 for the TNT and TEMP models on the
Wikidata 12k dataset. This is because they are not required to
perform negative sampling during training and thus the additional
time consumption only comes from the filtering and scoring. HyTE and
DE require negative sampling. However, since the size of the hard
negatives only increases sub-linearly with the size of the input
TKG, the additional time consumption of negative sampling does not
increase significantly for large TKG (i.e., ICEWS 05-15). This
demonstrates that \textit{Booster} can improve the performance of
existing TKGC models with an acceptable time consumption.

\textbf{Throughput w.r.t. hyper-parameters.}
Figure~\ref{fig:efficency} shows the throughput of filtering
strategies and the hierarchical scoring algorithm with varying
hyper-parameters. First, our proposed strategies achieve high
throughput to all the hyper-parameters. The average processing time
is nearly 0.5 ms per sample. Second, the throughput of our proposed
strategies decreases sub-linearly when the span of the time window
increases. This is because of the locality of TKGs that facts are
mostly short-term related. Our strategies meet this property and
thus can effectively filter out useless samples.

\textbf{Tradeoff between MRR and training time.} We conduct
experiments to investigate the tradeoff between $MRR$ and training
time. Specifically, we set $L_r$ and $L_t$ as 1, 3, and 5. By
varying $L_e$, we report the tradeoff between $MRR$ and training
time in Figure~\ref{fig:MRR_scal_STD_Time}(a). We observe that when
$L_r$ and $L_t$ are small, the MRR result increases as the training
time increases. When $L_r = L_t = 5$ and $L_e$ reaches 10, the
training time keeps increasing but MRR degrades drastically.

\textbf{Memory, CPU, and GPU usages.} We use
Psutil\footnote{https://pypi.org/project/psutil/} to keep track of
the memory and CPU usages and
GPUtil\footnote{https://pypi.org/project/GPUtil/} to collect the GPU
usage. The maximum memory usage of \textit{Booster} is 14.51 GB. The
total CPU utilization is 425\% (the CPU has 10 cores and full CPU
utilization is 1000\%). The GPU utilization rate is 34\%. As shown
in Figure~\ref{fig:MRR_scal_STD_Time}(b), the memory usage and
processing time of our algorithm increase sub-linearly with
increasing input graph size.

\begin{table}[t] \footnotesize
\centering \caption{Performance comparison on the ICEWS 14 dataset
with different sparsities of timestamps and entities.}
\scalebox{0.7}{
\begin{tabular}{c|c|cc|cc|cc}
\hline \multicolumn{2}{c|}{\textbf{Models}}                   &
\multicolumn{1}{c}{\textbf{HyTE}} &
\multicolumn{1}{c|}{\textbf{HyTE$_\textit{Booster}$}} &
\multicolumn{1}{c}{\textbf{TA}} & \multicolumn{1}{c|}{\textbf{TA$
_\textit{Booster}$}} & \multicolumn{1}{c}{\textbf{TNT}} &
\multicolumn{1}{c}{\textbf{TNT$_\textit{Booster}$}}
\\ \hline \multicolumn{1}{c|}{{}} & \multicolumn{1}{c|}{{Scope}}
& MRR         & MRR  & MRR         & MRR & MRR & MRR            \\
\hline & [0:100] & 0.255     & 0.291 & 0.421     & 0.447 & 0.598 &
0.621
\\ Time &[100:250] &0.276  & 0.315 &0.388         & 0.398 & 0.601            & 0.602
\\ &[300:] &0.306  & 0.309
& 0.387         & 0.388 & 0.623            & 0.640
\\ \hline &[0:10] & 0.149 &
0.198        & 0.324       & 0.361 & 0.403 & 0.421
\\ Entity &[10:50] & 0.287 &
0.306        & 0.378       & 0.383 & 0.585
& 0.598 \\
 &[100:] & 0.343 & 0.357
& 0.411         & 0.428 & 0.634            & 0.639
\\ \hline
\end{tabular}}
\label{tab:different sparsity}
\end{table}

\subsection{Balance and Stability (RQ4)}

\textbf{Balance.} We analyze the performance improvement of
\textit{Booster} for samples with varying sparsity in
Table~\ref{tab:different sparsity}. We first split the test samples
based on the scope of their time sparsity (number of facts in each
timestamp) and entity sparsity (number of interacted entities), and
calculate the average $MRR$ of samples in each scope. We can see
that \textit{Booster} can achieve performance improvement for
samples with different sparsities. In particular, the improvement is
larger for sparser samples, e.g., for samples with the number of
interacted entities in the scope of [0:10], \textit{Booster}
achieves 11.7\% improvement for TA, larger than 4.1\% improvement
for samples with the number of interacted entities in the scope of
[100:]. This observation shows that \textit{Booster} can effectively
enrich the sparse graph structures and thus achieve more balanced
performance across timestamps and entities. As shown in
Figure~\ref{fig:MRR_scal_STD_Time}(c) and (d), we provide the
standard deviation of the average $rank$ metric among timestamps. We
can see that by applying the \textit{Booster} framework, baseline
models can achieve lower standard deviations on different datasets,
which also verifies the effectiveness of our framework in reducing
the performance imbalance across timestamps.

\begin{figure}[t]
\centering \scalebox{0.95}{
\subfigure[]{\includegraphics[width=0.5\linewidth]{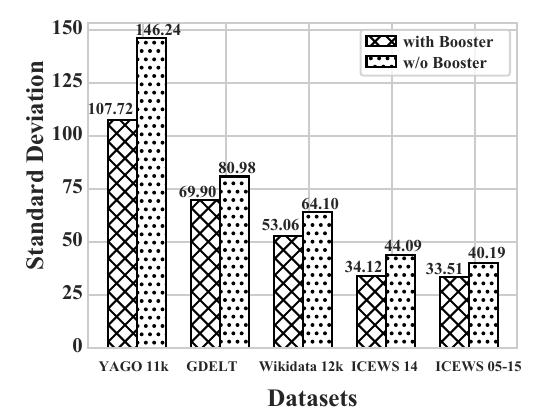}}
\subfigure[]{\includegraphics[width=0.5\linewidth]{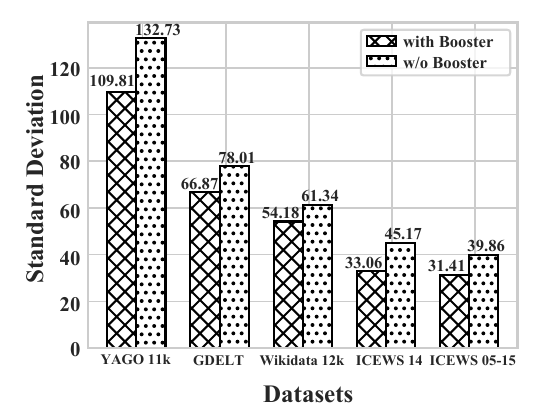}}}
\caption{(a) Statistical study on the variance reduction of the TEMP
model. (b) Statistical study on the variance reduction of the TNT
model.} \label{fig:variance_reduction}
\end{figure}

\textbf{Stability.} To verify the effectiveness of \textit{Booster}
on stabilizing the performance, we randomly select four test samples
whose performance results gradually get worse during training, and
then we show the change of their $rank$ metrics to training epochs
on the TEMP model. As illustrated in Figure~\ref{fig:case_example},
all of these samples achieve better performance with a smaller
fluctuation range when trained by the \textit{Booster} framework.
This is because \textit{Booster} can alleviate the misleading of
model preferences and avoid false negatives. In
Figure~\ref{fig:variance_reduction}, we show the standard deviations
of $rank$ of four independent training procedures. We can see that
for all datasets, \textit{Booster} can help different baseline
models to get smaller standard deviations, which verifies its
effectiveness in stabilizing performance.

\section{Conclusion}

In this paper, we make the first attempt to tackle the imbalanced
data and model preference issues for temporal knowledge graph
completion. We experimentally demonstrate existing methods'
limitations, and then propose the first pattern-aware data
augmentation framework tailored to TKGs to mitigate the impact of
imbalanced data and model preferences. Extensive experiments on five
datasets demonstrate that \textit{Booster} can help existing models
achieve higher performance with better balance. For our future work,
one promising direction is to study the robustness of existing TKG
completion methods.

\bibliographystyle{ACM-Reference-Format}
\bibliography{sample}

\end{document}